\documentclass[a4paper,12pt,letterpaper,headsepline,singlespacing,headsepline, french,oneside]{report}

\usepackage{geometry}
\usepackage[toc,page]{appendix} 
\usepackage[utf8]{inputenc}
\usepackage{geometry}
\usepackage[export]{adjustbox}
\usepackage{wrapfig}
\usepackage{float}
\usepackage{wrapfig}
\usepackage[T1]{fontenc}
\usepackage{hyperref}
\usepackage{caption}
\usepackage{indentfirst}
\usepackage[autolanguage]{numprint}
\usepackage{array,multirow,makecell}
\usepackage[table]{xcolor}
\usepackage{graphicx}
\usepackage{lmodern}
\usepackage{amsmath}
\usepackage{amssymb}
\usepackage{mathrsfs}
\usepackage{longtable}
\usepackage{array}
\usepackage[counterclockwise]{rotating}
\usepackage{multirow}
\usepackage[linesnumbered,ruled]{algorithm2e}
\usepackage{algorithm2e}
\usepackage{rotating}
\usepackage{lipsum}
\usepackage{hyperref}
\usepackage{footnote}

\usepackage{color}

\makesavenoteenv{table}
\usepackage{fancyhdr} 
\usepackage{titlesec}
\usepackage{amsmath}
\usepackage[francais]{babel}  
\newcolumntype{R}[1]{>{\raggedleft\arraybackslash }b{#1}}
\newcolumntype{L}[1]{>{\raggedright\arraybackslash }b{#1}}
\newcolumntype{C}[1]{>{\centering\arraybackslash }b{#1}}
\usepackage{titlesec}

\begin{document}

\begin{titlepage}
\center
\textbf{UNIVERSITÉ DE TUNIS EL MANAR} \\
\vspace{0.1em}
\textbf{FACULTÉ DES SCIENCES MATHÉMATIQUES, PHYSIQUES ET NATURELLES DE TUNIS}
\vspace{1em}
\begin{figure}[!ht]
\centering
\includegraphics[scale=0.13]{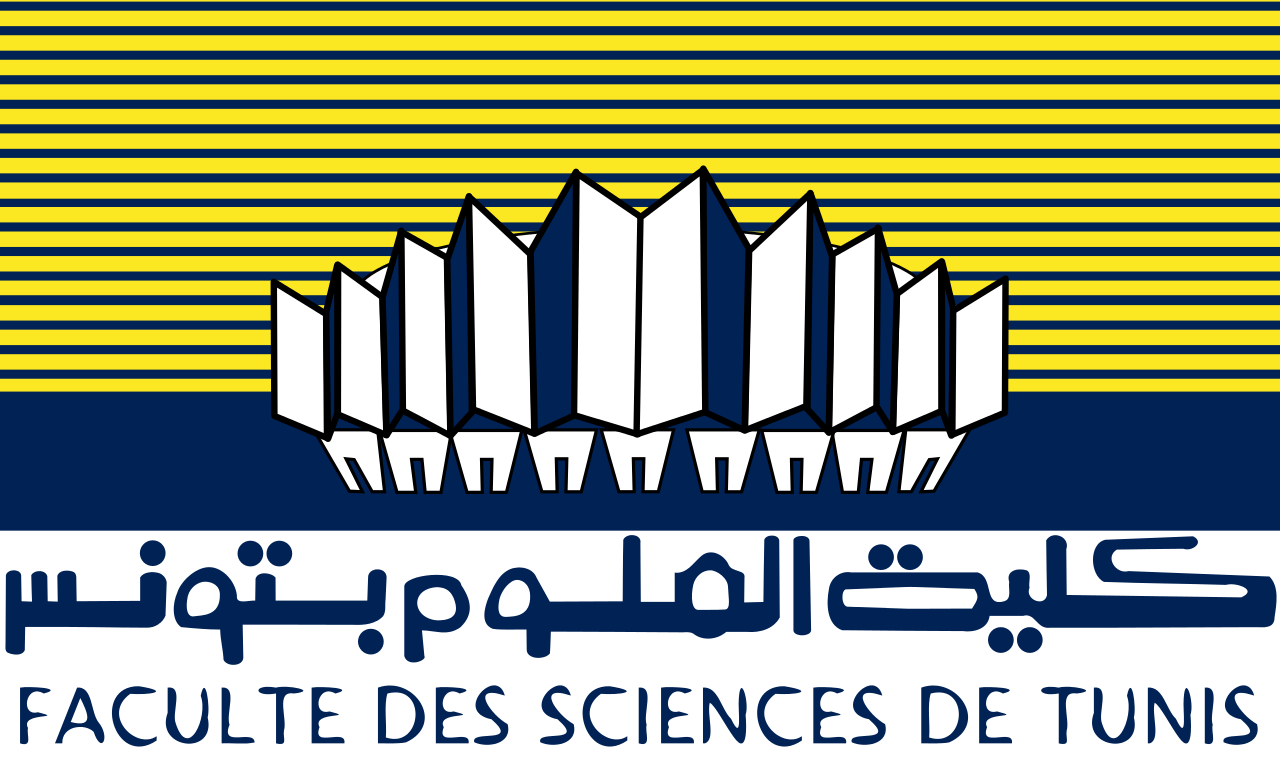}
\end{figure}
\vspace{2.3em} \\
{\bfseries \Large Mémoire de Mastère } \\
\vspace{0.1em}
{\slshape \normalsize Présenté en vue de l’obtention du diplôme de } \\
\vspace{1em}
{\textrm {\large \textbf{MASTÈRE DE RECHERCHE EN INFORMATIQUE}}} \\

{\slshape \normalsize Par } \\
\vspace{0.1em}
{\bfseries \large 	Ons KHEMIRI} \\
\vspace{3em}
{\textrm {\huge \textbf{Proposition d'une nouvelle approche d'extraction des motifs fermés fréquents}}} \\
\vspace{5em}

\textbf{Soutenue le 19/02/2018  devant le jury composé de :} \\

\begin{flushleft}
\textbf{\footnotesize  M.Mohamed Mohsen GAMMOUDI} \hspace{4ex}\textbf{\textit{{\footnotesize  Professeur à l’ISAMM}}} \hspace{8ex}\textbf{\textit{{\footnotesize Président}}} \\
\textbf{{\footnotesize  M.Samir  ELLOUMI}}\hspace{20ex}\textbf{\textit{{\footnotesize Maître-Assistant à la FST}}} \hspace{5ex}\textbf{\textit{{\footnotesize Rapporteur}}} \\
\textbf{{\footnotesize  M.Sadok BEN YAHIA}}\hspace{17ex}\textbf{\textit{{\footnotesize  Professeur à la FST}}} \hspace{11ex}\textbf{\textit{{\footnotesize  Directeur de mémoire}}}\\
\textbf{{\footnotesize  Mme.Souad BOUASKER}}\hspace{14ex}\textbf{\textit{{\footnotesize Assistante à FSJEGJ  }}} \hspace{9ex}\textbf{\textit{{\footnotesize  Co-encadrante}}} 

\end{flushleft}\vspace{5mm}
\textbf{ Au sein du laboratoire LIPAH : FST }
\pagenumbering{gobble}
\end{titlepage}

\chapter*{Remerciements}
\thispagestyle{empty}
C'est un grand plaisir que je réserve cette page pour exprimer toute ma gratitude et ma reconnaissance à tous ceux qui m'ont aidé à la réalisation de ce travail qui m'a permis de m'épanouir professionnellement.\\
\\
Je tiens d'abord à adresser mes plus sincères remerciements et toute ma reconnaissance, à mon directeur de mémoire, \textsl{\textbf{M. Sadok BEN YAHYA}}, Professeur à la Faculté des Sciences de Tunis,Université de Tunis Elmanar, qui a toujours été présent à mes côtés pour m'orienter. Il m'a permis d'approfondir au maximum mes travaux afin de pouvoir être fier aujourd'hui du travail réalisé. \\
\\
Ce travail n'aurait pas vu le jour sans  l'aide et le constant encouragement de \textsl{ \textbf{Mme. Souad BOUASKER}}. A son contact, j'ai beaucoup profité de sa finesse d'analyse, tant lors de l'élaboration de réflexions scientifiques que pour la gestion des aspects politiques inhérents à toute activité de recherche, ainsi que pour son suivi conjoint et sa disponibilité.\\
 \\
 Par la même occasion, je remercie tout spécialement l'ensemble du personnel et des enseignants de la Faculté des Sciences de Tunis et plus précisément au département de l'informatique pour leur implication pendant toutes mes années de formation. C'est grâce à qui j'ai acquis de précieuses connaissances lors de mon passage à la FST.\\
\\
Mes remerciements sincères s'adressent aux membres de jury pour l'honneur qu'ils me font d'accepter de juger mon travail.\\

\chapter*{Dédicace}





\begin{center}
\textrm{\textbf{À  mes parents}}
\end{center}
\begin{center}
Aucun hommage ne pourrait être à la hauteur de l’amour dont ils ne cessent de me combler. Que dieu leur procure bonne santé et longue vie
\end{center}

\begin{center}
\textrm{\textbf{  À ma très chère sœur Oumaima}}
\end{center}
\begin{center}
En souvenir d’une enfance dont nous avons partagé les meilleurs et les plus agréables moments.
\end{center}

\begin{center}
\textrm{\textbf{   À mon cher petit frère Taher}}
\end{center}
\begin{center}
Pour toute l’ambiance dont tu m’as entouré, pour toute la spontanéité et ton élan chaleureux, Je te dédie ce travail .
\end{center}

À qui je souhaite toute la réussite et le bonheur.
À toute personne qui m'a soutenue durant la période de ce projet.
À tous ceux qui m'ont soutenu de près ou de loin à réaliser ce travail.
Ainsi que tous ceux dont je n'ai pas indiqué le nom mais qui ne sont pas moins chers. 
 \newpage
\pagenumbering{arabic}
\tableofcontents

\listoffigures
\listoftables

\chapter*{Introduction générale}
\addcontentsline{toc}{chapter}{Introduction générale}
\markboth{INTRODUCTION GÉNÉRALE}{}

\begin{figure}[H]
\centering
\includegraphics[scale=0.8]{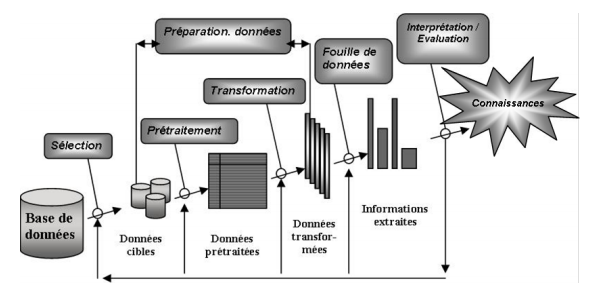}
\caption{  Étapes du processus de l’ECD }

\label{figetcd}
\end{figure}

La notion de  "Knowledge Discovery from Databases KDD" ou en Français "Extraction de Connaissances à partir de Données ECD", a été initialement introduite au début des années 1990 \cite{piateski1991knowledge,frawley1992knowledge}.  L’Extraction des Connaissances à partir de Données  est une discipline qui regroupe tous les domaines des bases de données, des statistiques, de l’intelligence artificielle et de l’interface homme-machine \cite{fayyad1996data}.\\
Le principal objectif de cette discipline étant de découvrir des connaissances nouvelles, pertinentes et cachées par le volume important de données. La réalisation de cet objectif nécessite la conception et la mise au point de méthodes permettant d’extraire les informations essentielles et cachées qui seront interprétées par les experts afin de les transformer en connaissances utiles à l’aide à la décision. \\
Fayyad \cite{kurgan2006survey} décrit le processus d’extraction de connaissances à partir de bases de données comme un processus itératif composé de plusieurs étapes. Ce processus suscite un fort intérêt industriel, notamment pour son champ d’application très large, pour son coût de mise en œuvre relativement faible, et surtout pour l’aide qu’il peut apporter à la prise de décision. 
Le processus d’ECD peut être découpé en cinq grandes étapes comme l’illustre  la Figure \ref{figetcd}. Ce processus commence par sélectionner le sous-ensemble des données qui peuvent être effectivement intéressantes. Vient ensuite l’étape de pré traitement visant quant à elle à corriger les données manquantes ou erronées. Puis, il faut transformer les données pour qu’elles soient utilisables par l’algorithme de choix. Celui-ci génère un certain nombre de motifs qu’il faut interpréter pour enfin obtenir des nouvelles connaissances. 

L'étape principale dans le processus de ECD est l’étape de fouille de données. C'est la partie la plus complexe du point de vue algorithmique. De nombreuses méthodes
existent alliant statistiques, mathématiques et informatique, de la régression linéaire à l’extraction de motifs fréquents.
L’un des objectifs fréquemment recherchés en fouille de données est la facilité  d'interpréter les connaissances extraites.

 


\section*{Contexte et problématique }

Notre mémoire de mastère s’inscrit dans le cadre du traitement des données. Dans ce travail,  nous nous intéressons à la fouille des motifs fermés fréquents à partir des bases de données . En effet, l'augmentation  du volume de données a donné naissance à diverses  problématiques liées à la collecte, au stockage, à l’analyse et l’exploitation de ces données pour créer de la valeur ajoutée. 
Pour traiter des données, une solution consiste à partitionner l’espace de recherche en des sous-contextes, puis explorer ces derniers simultanément. Dans le cadre de nos travaux de recherche, nous  proposons une approche  visant  l’extraction des motifs fermés fréquents, ainsi que leurs générateurs minimaux associés à partir des bases des transactions. En effet, l’idée principale de notre approche est de mettre  à jour des motifs fermés fréquents avec  leurs générateurs minimaux, et ce, en appliquant une stratégie de partitionnement du contexte d’extraction initial.
\section*{Organisation du mémoire }

Les résultats de nos travaux de recherche sont synthétisés dans ce mémoire qui est composé de quatre chapitres. 
\begin{itemize}
    \item {Le \textbf{premier chapitre} introduit les notions de base qui seront utilisées tout au long de ce travail. Ces dernières   incluent les notions préliminaires relatives à l'extraction des motifs fréquents et à l’analyse des concepts formels AFC.}
    \item{Le \textbf{deuxième chapitre} décrit l’état de l’art des approches séquentielles d'extraction des motifs fréquents ainsi que les motifs fermés fréquents. Nous y étudions et analysons aussi les approches de fouille parallèle.}
    \item{Dans \textbf{ le troisième chapitre}, nous introduisons notre nouvelle approche UFCIGs-DAC d’extraction des motifs fermés fréquents, ainsi que leurs générateurs minimaux. Nous allons présenter  la démarche et le processus de déroulement de notre approche.}
    \item{Le \textbf{quatrième chapitre} présente les études expérimentales menées sur des bases de test Benchmark. Cette étude s'étalera sur deux principaux axes. Le premier axe concerne la comparaison du temps de réponse de notre algorithme. Le second axe, concerne le nombre des motifs fermés fréquents extraits par notre approche  UFCIGs-DAC.}
\end{itemize}
Pour finir, nous clôturons ce mémoire par une conclusions générale récapitulant les résultats de nos travaux ainsi qu'un nombre de perspectives futures de recherches.

\chapter{ Notions de base }

\section{Introduction}

Ce chapitre a pour objectif de définir les notions de bases, qui constituent le point de départ dans la présentation de notre approche. De ce point de vue, la première section sera consacrée aux concepts nécessaires utiles dans l'extraction des motifs.
Nous enchaînons dans, la deuxième section avec la présentation de la méthode d'analyse des concepts formels $\mathit{ACF}$.

   \section{Extraction des motifs}

Nous commençons par définir l'ensemble des notions de base relatives à la technique  d'extraction des motifs, qui seront utilisés tout au long de ce travail. Définissons d'abord la  version de base d'extraction de motifs qui permet de faire la fouille dans une base des transactions.
    
 \subsection*{Base de transactions }
Une base de transactions (appelée aussi contexte d'extraction) est défini par un triplet  $\emph{D} = (\mathit{T , I, R})$  ou  :

\begin{itemize}
    \item {$T$:est un ensemble fini de transactions (ou objets).}
    \item{$\mathit{I}$ :est un ensemble fini d'items (ou attributs).}
    \item{$\mathit{R}$ : est une relation binaire $\mathit{R}$ \begin{math}\subseteq \end{math} $ \mathit{T}$ $\ast$ $\mathit{I}$ entre les transactions et les items.}
    
\end{itemize}

 Un couple (t, i) \begin{math}  \in \end{math}    $\mathit{R}$   dénote le fait que la transaction $t \in  \mathit{T}$ contient l'item   $i  \in  \mathit{I} $.
\\

\textbf{Exemple :} Un exemple d'une base de transactions  $\emph{D} = (\mathit{T , I, R})$  (resp contexte d'extraction  $\mathit{K} = (\mathit{O , I, R})) $ est donné par la table \ref{tab1.1}. Dans cette base (resp. ce contexte), l'ensemble de transactions $\mathit{T=\{1, 2, 3, 4, 5\} }$
(resp. d'objets $ O = \{1, 2, 3, 4, 5\})$ et l'ensemble
d'items $ I = \{A, B, C, D, E\}$. Le couple $(2, B) \in R$ car la transaction  $2 \in T $ contient l'item $ B \in I $.

\begin{center}

    \begin{tabular}{||p{2cm}||p{4cm}||}
        \hline  \hline    
   \rowcolor[gray]{.9}     
  TID    &   Items\\
         \hline  \hline  
1 & ACD\\  
\hline \hline 
2 & BCE\\
\hline \hline 
3 & ABCE\\
\hline \hline 
4 & BE \\
\hline  \hline 
5 & ABCE \\
\hline \hline 
6 & BCE \\
\hline \hline

\end{tabular} 
\captionof{table}{  (Base de Transactions)   $\mathit{D} , T = \{1, 2, 3, 4, 5, 6\} \ et \ I = \{A, B, C, D, E\}$}
\label{tab1.1}
    \end{center}
Le tableau ci-dessous représente également la base de transactions  $\mathit{D}$ en mode binaire.

\begin{center}

    \begin{tabular}{||p{2cm}|| p{1cm}p{1cm}p{1cm}p{1cm}p{1cm}||}
        \hline  \hline     
  \rowcolor[gray]{.9}     
  TID& A  &B & C & D&  E\\
         \hline   \hline  
1 & 1 &  0 & 1 & 1 & 0\\  

2 & 0& 1& 1& 0 &1\\
 
3 & 1& 1& 1& 0 &1\\

4 & 0& 1& 0& 0 &1 \\
 
5 & 1& 1& 1& 0 &1 \\

6 & 0& 1& 1& 0 &1 \\

\hline
\end{tabular} 
\captionof{table}{ Représentation en mode binaire de la base de transactions $\mathit{D}$}
    \end{center}
Nous notons, par souci de précision, que les notations de base de transactions et de contexte d'extraction seront les mêmes dans la suite. Ils seront notés $\emph{D} = (\mathit{T , I, R})$.

\subsection*{Motif ou Itemset}

Un motif, aussi appelé itemset, est un sous-ensemble non
vide de $\mathit{I}$ où  $\mathit{I}$ représente l’ensemble des items.
Une transaction t  $\in  \mathit{T}$ , avec un identificateur communément noté TID (Tuple IDentifier), contient un ensemble, non vide, d'items de $\mathit{I}$.
Un sous-ensemble I de $ \mathit{I}$ où k =$\left | I \right | $est
appelé un k-motif ou simplement un motif, et k représente la cardinalité de I Le nombre de transactions t d'une base $\mathit{D}$ contenant un motif $\mathit{I}$, $\left | \left \{ t\in D\mid I\subseteq t \right \} \right |$, est appelé support absolu de I et noté par la suite Supp ($\wedge I$) .

\subsection*{Treillis des motifs }
Un treillis des motifs\cite{stumme2002computing} est un regroupement conceptuel et hiérarchique des motifs. Il est aussi dit treillis d'inclusion ensembliste. Toutefois l'ensemble des parties de $ \mathit{I}$ est ordonné par inclusion ensembliste dans le treillis des motifs. Le treillis des motifs associé au contexte donné par le Tableau \ref{tab1.1} est représenté par la figure \ref{fig1.1}.
Toutefois, plusieurs mesures sont utilisées pour évaluer l'intérêt d'un motif, dont les plus connues sont présentées à travers la définition 4.

Dans la base des transactions illustrée dans le Tableau \ref{tab1.1},  $ t_1$ possède les motifs :$\phi$ , a, c, d, ac, ad, cd et acd. Parmi
l’ensemble global de $2^{P}$ motifs, on va chercher ceux qui apparaissent fréquemment. Pour
cela, on introduira les notions de connexion de Galois et de support d’un motif.

\subsection*{Supports d'un motif}
Le support d’un motif $\mathit{i}$ est donc le rapport de la cardinalité de l’ensemble des transactions qui contiennent tous les items de $\emph{I}$ par la cardinalité de l’ensemble de toutes les transactions.
Il capture la portée du motif, en mesurant sa fréquence d’occurrence.

support$(\textit{I})=\frac{{}\mid \psi(\textit{I})\mid }{\mid(\textsl{O})\mid}$.

Nous distinguons trois types de supports correspondants à $\mathit{I} $:
\begin{enumerate}

\item {\textbf{Support conjonctif :} $Supp(\wedge \textit{I} )=\mid\{t\in T\mid \forall i \in \textit{I} :(t,i) \in \mathfrak {R}\}\mid$}

\item{\textbf{Support disjonctif :} $Supp(\vee \textit{I} )=\mid\{t\in T\mid \exists  i \in \textit{I} :(t,i) \in \mathfrak R\}\mid$}

\item{\textbf{Support négatif :} $Supp(\neg \textit{I} )=\mid\{t\in T\mid \forall  i \in \textit{I} :(t,i) \notin \mathfrak R\}\mid .$}
\end{enumerate}

\textbf{Exemple :} Dans la base des transactions $D$, on trouve, Supp(A)= \begin{math} \frac{\left | \{1,3,5\} \right |}{6}=\frac{1}{2} \end{math}, et Supp(CE) = \begin{math}\frac{\left | \{2,3,5,6\} \right |}{6}=\frac{2}{3}\end{math}.
La valeur du support est décroissante, c'est à dire,  si $\mathit{I_1}$ est  un sous motif $\mathit{I} $ \begin{math}( \mathit{I_1} \supseteq   \mathit{I}) \end{math}
alors Support $(\mathit{I_1})$ \begin{math} \geq \end{math} Support$(\mathit{I})$.
Le support mesure la fréquence d’un motif : plus il est élevé, plus le motif est fréquent. On distinguera les motifs fréquents des motifs  in-fréquents à l’aide d’un seuil  minimal de support  conjonctif Minsupp. Dans la suite, s'il n'y a pas de risque de confusion, le support conjonctif sera simplement appelé support.


\subsection*{Correspondance de Galois\cite{ganter1999contextual}}

Soit le contexte d'extraction   \begin{math}\mathfrak{\emph{D}} = (\mathfrak \{\emph{T} , I, R\}) \end{math}. Soit l’application \begin{math}\phi\end{math} de l’ensemble des parties de $\emph{O}$ (c’est-à-dire l’ensemble de tous les sous-ensembles de $\emph{O})$, noté par $P(\emph{O})$, dans l’ensemble des parties de $\emph{I}$, noté par $P(\emph{I})$. L’application \begin{math}\phi \end{math}associe à un ensemble d’objets O\begin{math} \subseteq \end{math} $\emph{O}$, l’ensemble des items i \begin{math}\in \emph{I}\end{math} communs à tous les objets o \begin{math}\in 
\emph{O}\end{math}.

    \begin{center}

  $$ \phi:\emph{P}(\emph{O})\rightarrow\emph{P}(\emph{I})$$
    \\
  $$ \phi(O) =\{i \in \emph{I}\mid\forall o\in O,(o,i)\in\emph{R}\}$$

    
\end{center}

Soit l’application \begin{math}\Psi \end{math} de l’ensemble des parties de \emph{I} dans l’ensemble des parties de $\emph{O}$.
Elle associe à un ensemble d’items $\mathit{I}$ \begin{math}\subseteq \end{math} $\emph{I}$, l’ensemble d’objets o \begin{math}\in \end{math} $ \emph{O}$ communs à tous les
items i \begin{math}\in \end{math} $\mathit{I}$ :

    \begin{center}

   $$ \Psi :\emph{P}(\emph{I})\rightarrow\emph{P}(\emph{O})$$
    \\
   $$ \phi(O) =\{o \in \emph{O}\mid\forall i\in \mathit{I},(o,i)\in\emph{R}\}$$
     
    
\end{center}

\begin{math}\phi \end{math}(\emph{O} ) dénote l’ensemble de tous les items communs à un groupe de transactions \emph{T} (intension),
et \begin{math} \psi \end{math}(\emph{I}) l’ensemble de toutes les transactions partageant les même items de \emph{I} (extension).

Le couple \begin{math}(\psi , \phi) \end{math}définit une correspondance de Galois entre \emph{I} et \emph{T}
Par exemple, dans la base de données du tableau 2.2, nous avons  \begin{math}\phi \end{math} $(\{4, 6\}) = \{B, E\}$
et  \begin{math}\psi \end{math}$(\{A, C\}) = \{1, 3, 5\}$. Ce qui signifie que l’ensemble de transactions \{4, 6\} possède en
commun l’ensemble d’attributs $ \{B, E\}$. De la même manière, l’ensemble d’attributs $\{A, C\}$
possède en commun l’ensemble de transactions $\{1, 3, 5\}$.

 La définition suivante présente le statut de fréquence d'un motif, fréquent
ou in-fréquent, étant donné un seuil minimal de support
 

\subsection*{Motif fréquent-rare\cite{agrawal1993mining}}
Soit une base de transactions $\mathit{D} = (\mathit{T , I, R})$, un
seuil minimal de support conjonctif minsupp, un motif I  \begin{math}\subseteq \end{math} $ \mathit{I}$ est dit fréquent  si Supp(\begin{math}\wedge\end{math}$\mathfrak{I} $)
 \begin{math} \geq  \end{math} minsupp. I est dit infréquent ou rare sinon.
\\
\textbf{ Exemple 3 } Dans le tableau \ref{tab1.1}, en fixant une valeur  minsupp = \begin{math} \frac{3}{6}\end{math}, on obtient que $ \{A\}$ et $\{BC\}$ sont fréquents, alors que  le motif $\{ABC\}$ est in-fréquent ou rare .

\section{Analyse de concepts formels}

L'analyse de concepts formels (ACF), est une  méthode mathématique de classification, a été introduite dans \cite{barbut1970ordre},cette approche a été popularisée par Wille qui a utilisé le treillis de Galois\cite{tekaya2005algorithme} comme base de l'ACF \cite{wille1982restructuring}. Cette méthode a pour objectif de découvrir et d'organiser hiérarchiquement es regroupement possibles d'éléments ayant des caractéristiques en communs.
La notion de treillis de Galois est utilisée comme base de l'ACF.Chaque élément du treillis est considéré comme un concept formel et le graphe(diagramme de Hasse) comme une relation de généralisation -spécialisation entre les concepts.

\subsection*{ Concept formel}

Soit$ (\mathit{T , I, R})$ un contexte formel. Un concept formel est un couple (A, B) tel que $ A \subseteq  T$,
$B \subseteq  I$, $A^{'} = B \ et B^{'} = A.$ A et B sont respectivement appelés extension (extent) et intension (intent) du concept formel (A, B).\\

Un motif fermé est l'intension d'un concept formel alors que son support est la cardinalité de l'extension du concept.


\subsection*{Classe d'équivalence}
  L'ensemble des parties de \emph{I} est divisé en des sous-ensembles distincts ,nommés  aussi classes d'équivalence \cite{bastide2000mining}.les éléments de chaque classe possèdent la même fermeture :
  
Soit  $I subste I$, la classe
d’équivalence de $I$, dénotée , $\left [ I \right ]$
, est :   $ \left [I \right ] = I_1 \subseteq I\mid \lambda(I) = \lambda(I_1)$.
Les éléments de la classe  d'équivalence  $\left [ I \right ]$ ont ainsi la même valeur de support.

\begin{figure}[H]
\centering
\includegraphics[scale=0.8]{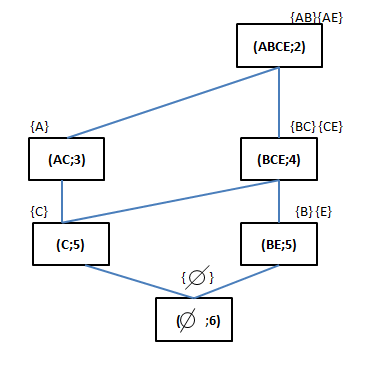}
\caption{ Le treillis d’Iceberg associé pour minsupp = 2. }

\label{figtr}
\end{figure}
 
\textbf{Exemple} : Étant donné le contexte d'extraction de tableau \ref{tab1.1},la figure \ref{figtr} représente Le treillis d’Iceberg associé pour minsupp = 2. Chaque nœud
dans ce treillis représente une classe d’équivalence. Il contient un motif fermé fréquent ainsi que
son support et est étiqueté par les générateurs minimaux associés.

Deux classes d’équivalence sont dites comparables si leurs motifs fermés associés peuvent être ordonnés par inclusion ensembliste, sinon elles sont dites incomparables.
La définition d’une classe d’équivalence nous amène à celle d’un générateur minimal.

\subsection*{Générateur minimal}
 Un itemset g  \begin{math} \subseteq \mathfrak{I} \end{math} est un générateur  minimal \cite{bastide2000mining,stumme2002computing} d’un itemset fermé  $\textit{I}$ si  et seulement si  $\lambda(g) = \textit{I}$ et \begin{math} \nexists \end{math} g' \begin{math} \subseteq  \end{math} g  tel  que $\lambda(g') = \textit{I}$ .

\textbf{Exemple : }
Dans la base $D$ du tableau \ref{tab1.1}, l'itemset $\{AB\}$ est un générateur minimal de
$\{ABCE\}$ puisque \begin{math}\lambda \end{math}$(\{AB\})$ = $\{ABCE\}$ et aucun de ses sous-ensembles propres a l'itemset $\{ABCE\}$ comme fermeture.

Ainsi, un motif fermé fréquent apparaît dans le même ensemble d’objets et par conséquent
il a le même support que celui de ses générateurs. Il représente donc un ensemble maximal
partageant les mêmes items, tandis que ses générateurs minimaux représentent les plus petits
éléments décrivant l’ensemble d’objets. 
Ainsi, tout motif est nécessairement compris entre un générateur minimal et un motif fermé.

Nous allons maintenant nous focaliser sur des propriétés structurelles importantes associées
à l’ensemble des motifs fermés et à l’ensemble des générateurs minimaux.
\subsection*{Treillis de concepts formels (de Galois)}

 Étant donné un contexte d’extraction K, l’ensemble
de concepts formels $ C_K$, extrait à partir de K, est un treillis complet $\mathfrak{ L_CK}$ =
$(C_K, \leq)$, appelé treillis de concepts (ou treillis de Galois), quand l’ensemble $C_K$ est considéré avec la relation d’inclusion ensembliste entre les motifs  \cite{ganter1999formal,barbutordre} : 
soient $ c_1 = (O_1, I_1) \ et c_2 = (O_2, I_2) $ deux concepts formels, $c_1 \leq c_2$ si $  I_1 \sqsubseteq  I_2  $.

Outre la contrainte de fréquence minimale traduite par le seuil minsupp, d'autres contraintes
peuvent être intégrées dans le processus d'extraction des motifs. Ces contraintes admettent
différents types, dont les deux principaux sont définis dans ce qui suit \cite{han2004sequential}.

\subsection*{Contrainte anti-monotone/monotone}
 Dans le processus d'extraction des motifs , deux contraintes principales sont\ définies dans  \cite{pei2002constrained} comme suit : 
\begin{itemize}
\item{Une contrainte  \emph{Q} est anti-monotone si  \begin{math}\forall \mathit{I} \subseteq \emph{I},\forall \mathit{I_1} \subseteq  \mathit{I}:\mathit{I_1} satisfait \emph{Q} \Rightarrow  \mathit{I} satisfait \emph{Q} \end{math}}

\item{Une contrainte  \emph{Q} est monotone si  \begin{math}\forall \mathit{I} \subseteq \emph{I},\forall \mathit{I_1} \supseteq  \mathit{I}:\mathit{I_1} satisfait \emph{Q} \Rightarrow  \mathit{I} satisfait \emph{Q} \end{math}}

\end{itemize}
\textbf{Exemple :}

Soit P ($\mathit{I}$) l'ensemble de tous les sous-ensembles de $\mathit{I}$. Dans ce qui suit, nous introduisons
les notions duales d'idéal d'ordre et de filtre d'ordre \cite{priss2006formal} définis sur P ($\mathit{I}$).

\subsection*{Idéal d'ordre}
Soit  \emph{P}(\emph{I})  l'ensemble  de tous les sous-ensembles \emph{ S} de \emph{I} .La notion idéal d'ordre est introduit dans \cite{mineau1999conceptual} . 
\emph{S} de \emph{P}(\emph{I}) est un idéal d'ordre s'il vérifie les propriétés suivantes:
\begin{itemize}

\item  {Si  \begin{math}\mathit{I} \in \emph{S} , alors  \forall \mathit{I_1} \subseteq  \mathit{I} :\mathit{I_1} \in \emph{S}\end{math}}

\item  {Si  \begin{math}\mathit{I} \notin \emph{S} , alors  \forall \mathit{I} \subseteq  \mathit{I_1} :\mathit{I_1} \notin \emph{S}\end{math}}

\end{itemize}

\subsection*{Filtre d'ordre}
Soit  \emph{P}(\emph{I})  l'ensemble de tous les sous-ensembles \emph{S} de \emph{I} .La notion filtre d'ordre est introduit dans \cite{mineau1999conceptual} . 
 \emph{S} de \emph{P}(\emph{I}) est un filtre d'ordre s'il vérifie les propriétés suivantes:

\begin{itemize}
    \item {Si  \begin{math}\mathit{I} \in \emph{S} , alors \  \forall \mathit{I_1} \supseteq   \mathit{I} :\mathit{I_1} \in \emph{S}\end{math}}

\item  {Si  \begin{math}\mathit{I} \notin \emph{S} , alors  \  \forall \mathit{I} \supseteq  \mathit{I_1} :\mathit{I_1} \notin \emph{S}\end{math}}

\end{itemize}

Une contrainte anti-monotone telle que la contrainte de fréquence induit un idéal d'ordre.
D'une manière duale, une contrainte monotone telle que la contrainte de rareté forme un
filtre d'ordre. L'ensemble des motifs satisfaisant une contrainte donnée est appelé théorie
dans \cite{mannila1997levelwise}. Cette théorie est délimitée par deux bordures, une dite
bordure positive et l'autre appelée bordure négative, et qui sont définies comme suit.

\subsection*{Bordure positive}
Compte tenu d'un seuil minimum de support minsupp.La bordure positive  \cite{mannila1997levelwise} \begin{math} Bd^{+}\end{math}  représentée par les motifs fréquents maximaux.
\begin{math} Bd^{+}\end{math}, est l’ensemble des plus grands itemsets fréquents (au sens de l’inclusion) dont tous les sur-ensembles sont fréquents, et est définie comme suit:
\\
\begin{center}

\begin{math} Bd^{+}\end{math}=  \begin{math}\{\mathit{I} \in \emph{I}\mid     supp(\mathit{I}) \geqslant minsupp, \forall \mathit{I_1} \supseteq  \mathit{I}, supp(\mathit{I_1}) \geqslant minsupp  \} \end{math}
\end{center}

\subsection*{Bordure négative}
Compte tenu d'un seuil minimum de support minsupp.
La bordure négative \cite{mannila1997levelwise}  \begin{math} Bd^{-}\end{math} est représentée  par les motifs in-fréquents minimaux.

\begin{math} Bd^{-}\end{math}, est l’ensemble des plus petits itemsets qui ne sont pas fréquents dont tous les sous ensembles
sont fréquents, et est définie comme suit :
\begin{center}

\begin{math} Bd^{-}\end{math}=  \begin{math}\{\mathit{I} \in \emph{I}\mid     supp(\mathit{I}) \leq minsupp, \forall \mathit{I_1} \subseteq  \mathit{I}, supp(\mathit{I_1}) \geqslant minsupp  \} \end{math}

\end{center}

\subsection*{Opérateur de fermeture}

Les applications  \begin{math}\lambda\end{math} = \begin{math}\Phi  \circ \Psi\end{math}
et \begin{math}\sigma = \Psi \circ \Phi \end{math} sont appelées les opérateurs de fermeture \cite{priss2006formal} de la correspondance de Galois \cite{ganter1999contextual}. \\

Par exemple, dans le contexte $\mathfrak{D}$ du tableau 2.2, on a si $T = \{3, 5\}$, alors \begin{math} \phi(T) \end{math} = $\{ABCE\}$ et donc \begin{math}\sigma \end{math} =$ \{3, 5\}$.
 Et si $T = \{1, 2, 3\}$, alors\begin{math}\phi(T) \end{math} = \{C\}
et donc \begin{math}\sigma \end{math} $= \{1, 2, 3, 5, 6\}$. 
Si $O = \{AC\}$, alors  \begin{math}\psi(O )\end{math} $= \{1, 3, 5\}$ et donc \begin{math}\lambda \end{math}$= \{AC\}$.
Dans ces exemples, les ensembles $ \{3, 5\}$ et $\{AC\}$ sont fermés.

L’opérateur de fermeture  \begin{math} \lambda \end{math}, tout comme  \begin{math}\sigma \end{math}, est caractérisé par le fait qu’il est :

\begin{enumerate}

\item{ Isotonie}.
\item{ Expansivité}
\item { Idempotence}.

\end{enumerate}

\subsubsection*{Types de motifs fréquents}
Nous allons maintenant introduire la notion de motif fréquent et celle de motif fermé.
Selon la nature des motifs fréquents nous pouvons  trouver deux types :

\subsubsection*{Motif fermé fréquent }
Un itemset \textit{I}  \begin{math}\sqsubseteq \end{math} \emph{I} est fermé si seulement si  $\textit{I}= \lambda(\textit{I}) \cite{pasquier1998pruning}$.\\
$\textit{I} $ est un ensemble maximal d’items communs à un ensemble d’objets \cite{pasquier1999efficient}. 
Un itemset fermé \textit{I} est fréquent si seulement si son support noté support(\textit{I}) \begin{math}=\frac{{}\mid \psi(\textit{I})\mid }{\mid(\textsl{O})\mid} \geq\end{math} minsup (i.e., le seuil minimal de support)
 \textbf{Exemple :}
 Dans la base de données B du tableau 2.2, les motifs $\{AB\}$, $\{ABC\}$ et $\{ABCE\}$ sont dans la même classe d’équivalence. Donc, $\{ABCE\}$ est l'itemset fermé.

 \subsubsection*{Motif fréquent maximal}
 Un motif fréquent est dit Maximal si aucun de ses sur-motifs immédiats n’est fréquent.
\\
\textbf{Exemple :} 
Le schéma suivant illustre la relation entre le motifs fréquents, fréquents fermés et fréquents maximaux :
 \begin{figure}[H]
\centering
\includegraphics[scale=0.8]{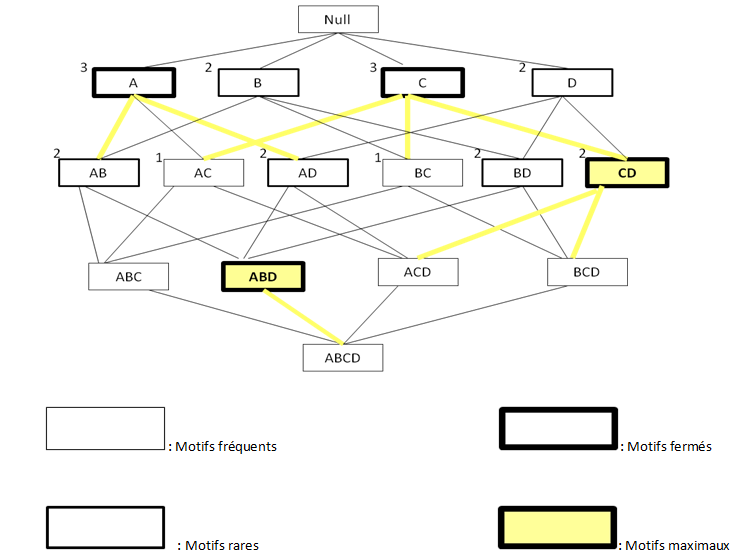}
\caption{Relation entre les motifs fréquents,fréquents fermés et fréquents maximaux  }

\label{fig1.1}
\end{figure}

 \begin{itemize}
     \item { Les motifs encadrés par les lignes minces ne sont pas fréquents, les autres le sont.}
     \item{ Les motifs encadrés par des lignes plus épais sont fermés.}
     \item{Les motifs  encadrés par des lignes plus épais et colorés  sont maximaux}
 \end{itemize}

 \begin{center}

$ \ Les \ motifs \ maximaux \sqsubset \  Les \ motifs \ fermés\sqsubset  \ Les \ motifs \ fréquents$

 \end{center}

\section{Conclusion }
Dans ce chapitre, nous avons présenté  l'ensemble des notions de bases relatives à l'extraction des motifs que nous utiliserons au chapitre suivant. De surcroît, nous nous avons focalisé sur les propriétés structurelles importantes à l'analyse des concepts formels. Le chapitre suivant est consacré à la présentation de l'état de l'art des approches traitant de l'extraction des motifs fréquents et des motifs fermés fréquents.

\chapter{État de l'art}
\section*{Introduction}

Étant donné que dans le cadre de ce mémoire, nous jugeons intéressant de consacrer ce chapitre à la présentation et l'étude des approches de l'état de l'art s'inscrivant dans le cadre de notre problématique. Dans ce chapitre, nous abordons dans la première section la présentation des approches d'extraction des motifs fréquents et des motifs fermés fréquents en séquentiel.
Dans la deuxième section nous présentons les approches d'extraction des motifs fréquents et des motifs fermés fréquents en parallèle. La troisième section sera  consacrée à une synthèse qui permet de classer les différentes approches d'extraction parallèle des motifs fréquents et des motifs fermés fréquents selon des différentes stratégies.

\section {Exploration séquentielle des motifs}

\subsection{Algorithmes d'extractions séquentielles des motifs fréquents}

Une approche naïve pour déterminer tous les motifs fréquents dans une base des données  $\emph{D}$, consiste simplement à déterminer le support (support) de toutes les combinaisons des items dans $\emph{D}$. Ensuite, garder seulement les items/motifs qui satisfont un seuil de support minimum (MinSup). 

Les algorithmes de recherche de motifs fréquents peuvent
se séparer en deux  stratégies:  en largeur d’abord \cite{agrawal1994fast},  ou en profondeur d'abord \cite{han2000mining}. 
\begin{itemize}
    \item {\textbf{Stratégie en largeur d'abord} \cite{agrawal1994fast} :En adoptant une stratégie en largeur d'abord , tous les k-itemsets candidats sont génères en faisant une auto-jointure de l'ensemble des $(k-1)$ itemsets fréquents: calculer les motifs du treillis en largeur d’abord, niveau par niveau. Le premier niveau, $ L_1$, est initialisé avec l’ensemble des items fréquents\cite{samet2014mining}. Chaque niveau $L_k$ est construit en combinant des motifs du niveau précédent, $L_k-1$.}
     \item {\textbf{Stratégie en profondeur d'abord} \cite{han2000mining}: Cette méthode consiste à énumérer les motifs fréquents dans un ordre prédéfini(par exemple ordre Lexicographique).}
    \item{\textbf{Stratégie hybride}\cite{hamrouni2008succinct}: explore en profondeur d'abord le treillis des itemsets, mais ne génère qu'un seul itemset à la fois.}
\end{itemize}

\subsubsection{Algorithme  \textsl{Apriori}}

L'algorithme  \textsl{Apriori} a été  proposé dans \cite{leskovec2014mining}.
Il se concentre sur la réduction de l'accès au disque d'E/S lors de l'extraction des motifs fréquents. \\
Pour ce faire, l'algorithme \textsl{Apriori} répond au critère d'anti-monotonie. C'est à dire, si un item/itemset n'est pas fréquent, alors tous ses sur-ensembles ne peuvent pas être fréquents.
 \textsl{Apriori} opère en deux étapes   pour extraire  les motifs  fréquents:
\begin{enumerate}
    \item {Étape de combinaison des items.}
    \item {Étape d'élagage.}
\end{enumerate}

Pour extraire les motifs fréquents,  \textsl{Apriori} parcourt la base de données $D$ et détermine une liste candidate $C_1$ d'items fréquents de taille 1, puis l'algorithme filtre $C_1$ et ne conserve que les items qui satisfont le support minimum $MinSup$ et les stocke dans une liste des fréquents $F_1$. À partir de $F_1$, l'algorithme génère les motifs candidats de taille 2 dans une liste disons $C_2$ et cela en combinant toutes les paires d'items fréquents de taille 1 en $F_1$.
Ensuite, \textsl{Apriori} analyse $D$ et détermine tous les motifs dans $C_2$ qui satisfont le support  $MinSup$, le résultat est stocké dans une liste $F_2$.
Le processus d'exploration d'Apriori est exécuté jusqu'à ce qu'il n'y ait plus  des motifs  candidats dans $D$ à vérifier.

\subsubsection{Algorithme \textsl{Eclat} }
L'algorithme \textsl{Eclat} a été introduit par Zaki dans \cite{zaki2000scalable}, consiste à effectuer un processus d'extraction des motifs fréquents  dans la mémoire sans accéder au disque. L'algorithme procède en stockant une liste d'identifiants de transaction (TID) dans la mémoire de chaque item de la base de données.
Pour déterminer le support d'un motif $I$, \textsl{Eclat} croise les TIDs de tous les items de $I$. Eclat effectue une recherche des itemsets fréquents en profondeur d’abord et se base sur le concept de classes d’équivalence. Par exemple, ABC et ABD appartiennent à la même classe d’équivalence. Deux k-itemsets appartiennent à une même classe d’équivalence s’ils ont en communun préfixe de taille (k -1).

\subsubsection{ Algorithme\textsl{FP-Growth}}
FP-Growth (Frequent-Pattern Growth)\cite{han2000mining}, a été considéré comme l'algorithme le plus performant par rapport aux autres algorithmes pour extraire des itemsets fréquents.

L'algorithme consiste d’abord à compresser la base de données en une structure compacte appelée \textit{FP-tree} (Frequent Pattern tree) et qui apporte une solution au problème de la fouille de motifs fréquents dans une grande base de données transactionnelle. Contrairement aux techniques mentionnées précédemment, l'algorithme \textsl{FP-Growth} ne repose sur aucune approche de génération de motifs candidats.

L’algorithme \textsl{FP-Growth} effectue deux passes (scans) à la base de transactions :
\begin{itemize}
    \item {\textbf{Passe 1} le premier passage de \textsl{FP-Growth} sur la base de données $\emph{D}$ est consacré à déterminer la valeur du  support de chaque item  dans $\emph{D}$. L'algorithme ne retient que les éléments fréquents dans une liste \textit{F-List}.
    Ensuite, \textsl{FP-Growth} trie \textit{F-List} dans un ordre décroissant en fonction du valeur de support et qui est  comparé avec le seuil de support préfixé (\textsl{MinSup}).}
    \item {\textbf{Passe 2}  un \textsl{FP-Tree} est construit par la création d’une racine vide et un second parcours de la base de données où chaque transaction est décrite dans l’ordre des items donné par la liste \textit{F-List}.
    
    Chaque nœud de l'arbre FP-Tree représente un élément dans L et chaque nœud est associé à un compteur (c'est-à-dire, compte de support initialisé à 1).
Si une transaction partage un préfixe commun avec une autre transaction, le compte de support de chaque nœud visité est incrémenté de 1. Pour faciliter la traversée de FP-Tree, une table d'en-tête est construite pour que chaque élément pointe vers ses occurrences dans l'arbre via une chaîne de liens-nœuds.
    
    }

\end{itemize}
En dernier lieu, le FP-Tree est fouillé par la création des (sub-)fragment conditionnels de base. En fait, pour trouver
ces fragments, on extrait pour chaque fragment de longueur 1 (suffix pattern) l’ensemble des préfixes existant dans
le chemin du FP-Tree (conditional pattern base). L’itemset fréquent est obtenu par la concaténation du suffixe avec
les fragments fréquents extraits des FP-Tree conditionnels.

\begin{center}

    \begin{tabular}{|p{1cm}|p{2cm}|}
        \hline \hline     
  TID    &   Items\\
         \hline     \hline 
1 & 1,2,5,6\\  
\hline   \hline 
2 & 2,4,7\\
\hline   \hline 
3 & 2,1,3,6\\
\hline   \hline 
4 & 6,2,4,7 \\
\hline   \hline

\end{tabular} 
\captionof{table}{ Exemple de base de transactions $\emph{D}$}
\label{tab}
    \end{center}

\begin{center}

    \begin{tabular}{|p{2cm}|p{2cm}|}
        \hline      
  Items  &   Support\\
         \hline  \hline    
2 & 4\\  
\hline \hline  
1 & 3\\
\hline \hline  
4 & 2\\
\hline \hline  
6 & 2 \\
\hline \hline  
7 & 2\\
\hline \hline  
3 & 1\\
\hline \hline  
5 & 1\\
\hline \hline

\end{tabular} 
\captionof{table}{  Items associés à leur support}
\label{tabb}
    \end{center}

\begin{figure}[H]
\centering
\includegraphics[scale=0.8]{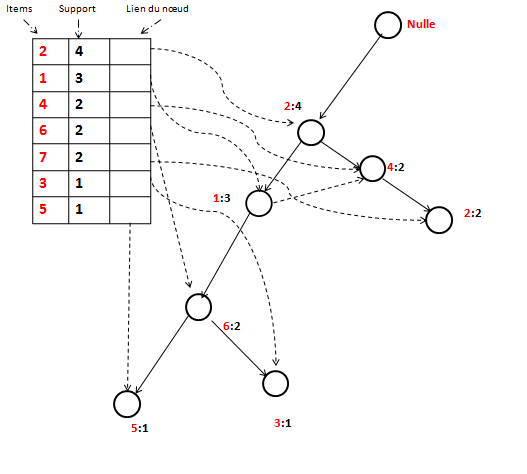}
\caption{Construction du FP-Tree}
\label{fig5}
\end{figure}
 En dernier lieu, le FP-Tree est fouillé par la création des (sub-)fragment conditionnels de base. En fait, pour trouver
ces fragments, on extrait pour chaque fragment de longueur 1 (suffix pattern) l’ensemble des préfixes existant dans
le chemin du FP-Tree (conditional pattern base). L’itemset fréquent est obtenu par la concaténation du suffixe avec
les fragments fréquents extraits des FP-Tree conditionnels (voir tableau \ref{tab2.3})

\begin{center}
\begin{tabular}{|p{2cm}|p{4cm}|p{4cm}|p{5cm}|}

        \hline       \hline
  Items  &   Motifs conditionnels &  FP-Tree conditionnels & Motifs fréquents\\
         \hline    \hline 
 5 & 1,2,6,2,1,3 & (2 : 2, 1 : 1) & 2, 5 : 2, 1, 5 : 2, 2, 1, 5 : 2\\
 \hline  \hline
  4 & 2 : 1, 2, 1 : 1  &(2 : 2)    & 2, 4 : 2\\
 
  \hline  \hline

\end{tabular}
\captionof{table}{  Fouille du FP-Tree}
\label{tab2.3}
\end{center}

\subsubsection{ Algorithme \textsl{SON} }
l'algorithme $\emph{SON}$ a été introduit dans \cite{savasere1995efficient}. Cet algorithme  permet l'extraction des itemsets fréquent.
Le principe d'extraction de $\emph{SON}$ est tiré du fait que l'ensemble de tous les itemsets fréquents globaux (c'est-à-dire tous les itemsets fréquents dans $\emph{D}$ est inclus dans l'union de l'ensemble de tous les itemsets fréquents locaux.
Pour déterminer l'ensemble des itemsets fréquents, le processus $\emph{SON}$ procède en effectuant un processus d'exploration en deux phases comme suit :
\begin{enumerate}
\item{\textbf{Phase1 :}\\ Diviser la base de données d'entrée $\emph{D}$ en n partitions de données,  $\emph{D}$\begin{math}=\{P_1\end{math},\begin{math}P_2\end{math},....,\begin{math}P_n\end{math}\} d'une manière que chaque \begin{math}P_i\end{math} dans $\emph{D}$ s'insère dans la mémoire disponible.Ensuite, extraire chaque partition de données \begin{math} P_i \end{math}dans la mémoire en fonction d'un support minimum local $\textsl{LMinSup}$ (le support minimum local est calculé en fonction du nombre de transactions en \begin{math} P_i \end{math}et du support minimum global donné  $\emph{GMinSup}$) et d'un algorithme $\emph{FIM}$ spécifique ( par exemple, algorithme Apriori ou une de ses améliorations). Ainsi, la première phase de l'algorithme $\emph{SON}$ est consacrée à la détermination d'une liste des itemsets  fréquents locaux $LF_I$.}

\item{\textbf{Phase2 :}\\ Cette phase passe en filtrant les itemsets fréquents locaux dans la liste $LF_I$ en fonction du support minimum global $\emph{GMinSup}$ .
Cette étape est effectuée pour valider la fréquence globale de l'ensemble des itemsets fréquents locaux.
L'algorithme $\emph{SON}$ analyse la base de données entière $\emph{D}$ et vérifie la fréquence de chaque ensemble d'items fréquents locaux dans $LF_I$.Ensuite, il renvoie une liste d'iemsets fréquents globaux ($GF_I$) qui est un sous-ensemble de $LF_I$, c'est-à-dire \begin{math} GF I \subseteq LF_I \end{math}. }

\end{enumerate}

\subsection{Discussion }

Dans cette sous-section, nous allons discuter les différents algorithmes d'extraction séquentielle des motifs fréquents.
Nous récapitulons dans le tableau \ref{tabif1} les caractéristiques des différentes approches étudiées.
Cette comparaison couvre les axes suivants :

\begin{itemize}
    \item{\textbf{ Stratégie d'exploration:}Cette propriété décrit la stratégie d'exploration des motifs fréquents\cite{ayouni2011extracting} .
    \item{\textbf{Caractéristiques :}Cette propriété décrit les caractéristiques de l'approche en question.}}
\end{itemize}
En comparant les algorithmes décrits dans cette sous section, nous pouvons noter  les remarques suivantes :
\begin{itemize}
    \item {\textbf{Apriori:} Malgré la propriété anti-monotonie, la performance de l'algorithme Apriori est proportionnelle à son nombre de candidats itemset à vérifier par rapport à la base de transactions. Une amélioration AprioriTID et Eclat \cite{agrawal1994fast}  qui consiste à intégrer les identificateurs des transactions (TIDs).}
   \item{ \textbf{FP-Growth :} L’avantage majeur de l’algorithme est qu’il ne fait que deux balayages de la base des transactions. De surcroît, il peut être considéré  comme un algorithme complet puisqu'il contient toutes les informations sur les éléments fréquents, ainsi es items sont classés en ordre de fréquence décroissante.Néanmoins, malgré sa structure compacte, cela ne garantit pas, dans le cas ou la base de transactions est trop volumineuse, que toute la structure du FP-tree tiendra en mémoire centrale.}
   \item{\textbf{SON} Comme il effectue deux analyses de bases de données, le système SON a montré de meilleures performances que l'algorithme Apriori. Cependant, la principale limitation de cet algorithme est sa première phase d'extraction. c'est-à-dire, dans le cas où une partition de données contient un grand nombre de MOTIFS fréquents locaux, dans ce cas, les performances de la seconde phase seraient également affectées.}

\end{itemize}

\newpage

    \begin{tabular}{||p{2cm}||p{3cm}||p{10cm}||}
        \hline    \hline  
  Algorithme    &Stratégie d'exploration  &\hspace{4em}Caractéristiques\\
         \hline  \hline  
    Apriori \cite{leskovec2014mining}& Largeur d'abord& \begin{itemize}
        \item { Complet:Pas de perte d'informations.}
        \item{Nombre considérable de l'accès à la base des transactions.}
        \item{Sa performance est proportionelle à son nombre de motifs candidats.}
    \end{itemize}.\\
    \hline\hline
    FP-Growth \cite{han2000mining}&Profondeur d'abord&\begin{itemize}
    \item {Deux balayages à la base des transactions.}
    \item{Complet}
    \item{A chaque éléments fréquent correspond un chemin dans l’arbre}
    \item{Structure compacte.}
\end{itemize}.\\
    \hline\hline
Eclat\cite{zaki2000scalable}&Profondeur d'abord&.\begin{itemize}
    \item {Deux balayages à la base des transactions.}
    \item{Rapide:compte les supports des items avec les TIDs.}
\end{itemize}\\
    \hline\hline
     SON  \cite{savasere1995efficient}&-&\begin{itemize}
         \item {Partitionnement de la bases des transactions.}
         \item{Utilise un algorithme FIM spécifique pour la fouille. }
         \item{Nombre considérable de l'accès à la base des transactions.}
     \end{itemize}.\\
     \hline\hline
     \end{tabular} 
 
\captionof{table}{Tableau comparatif des algorithmes d’extraction des motifs fréquents}
\label{tabif1}

        


  \newpage

 Afin de résoudre les problèmes rencontrés par les algorithmes d'extraction des itemsets fréquents, une nouvelle approche basée sur l'extraction des itemsets fermés fréquents est apparue\cite{yahia2006frequent}. Cette approche est basée sur la fermeture de la connexion de Galois\cite{nguifo2004fouille}. Elle est fondée sur un élagage du treillis des itemsets fermés, en utilisant les opérateurs de fermeture de la connexion de Galois. Plusieurs algorithmes ont été proposés dans la littérature, dont le but est de découvrir les itemsets fermés fréquents.

\subsection{Algorithmes d'extractions séquentielles des motifs fermés fréquents}

Dans la littérature, plusieurs algorithmes ont été proposés pour résoudre le problème d’extraction des motifs fréquents. En effet, ce problème a été d’abord introduit dans \cite{agrawal1994fast}.
Dans cette sous section, nous allons passer en revue les principaux algorithmes permettant l'extraction des motifs fermés fréquents en séquentiel.

les stratégies adoptées pour l'exploration de l'espace de recherche soient classées  en deux stratégies, à savoir la stratégie "Générer-et-tester", et la stratégie  "Diviser-et générer" 
\begin{itemize}
    \item {\textbf{La stratégie "Générer-et-tester"}\cite{hamrouni2010generalization}  : Les algorithmes adoptant cette stratégie parcourent l'espace de recherche par niveau. A chaque niveau k, un ensemble de candidats de taille k est génère. Cet ensemble de candidats est, généralement, élagué par la conjonction d'une métrique statistique ( le support) et des heuristiques basées essentiellement sur les propriétés structurelles des itemsets fermés et/ou des générateurs minimaux.\cite{ben2004approches}}
    \item{\textbf{La stratégie Diviser-et-régner} \cite{brahmi2012omc}: Les algorithmes adoptant cette stratégie essaient de diviser le contexte d'extraction en des sous-contextes et d'appliquer le processus de découverte des itemsets fermés récursivement sur ces sous-contextes. Ce processus de découverte repose sur un élagage du contexte base essentiellement sur l'utilisation d'une métrique statistique et d'heuristiques introduites\cite{ben2004approches}.}

\end{itemize}

\subsubsection{Algorithme \textsl{Close}}

L'algorithme \textsl{Close} a été proposé d'abord dans\cite{pasquier1999efficient}. C'est un algorithme itératif d'extraction des itemsets fermés fréquents, en parcourant l'ensemble des générateurs des itemsets fermés fréquents par niveaux.
Durant chaque itération k de l'algorithme, un ensemble \begin{math}FF_k\end{math} de k-générateurs candidats est considéré. Chaque élément de cet ensemble est constitué de trois champs :
\begin{enumerate}

\item{ k-générateur candidat.}
\item{La  fermeture de k-générateur, qui est un itemset fermé candidat.}
\item{Le support de k-générateur.}

\end{enumerate}

À la fin de l'itération k, l'algorithme stocke un ensemble \begin{math} FF_k \end{math} contenant les k-générateurs fréquents, leurs fermetures, qui sont des itemsets fermés fréquents, et leurs supports.\\


Ainsi, chaque itération est composée de deux étapes :
\begin{enumerate}
    \item {Etape d'élagage : dans cette étape, une fonction \textit{GEN-CLOSURE} est appliquée à chaque générateur de $FFC_k$,  déterminant ainsi son support et sa fermeture.}
    \item{Etape de construction: Après l'élagage des générateurs non fréquents, une fonction \textit{GEN-GENERATOR} utilise l'ensemble d'itemsets fermés fréquents $FF_k$ et calcule l’ensemble$ FFC_k+1$ contenant tous les (k + 1)-itemsets, qui seront utilisés dans l’itération suivante. À ce stade, l'ensemble $ FFC_k+1$ est élagué comme suit.  Pour tout $ c\in FFC_k+1$, si \textit{c} est inclus dans la fermeture d’un des sous-ensembles, i.e. les éléments de$ FC_k$ dont la jointure a permis d’obtenir \textit{c}. Dans ce cas, \textit{c} est éliminé $FFC_k+1$. L’algorithme s’arrête quand il n’y a plus de générateurs à traiter.
 }
\end{enumerate}

\begin{figure}[!ht]
\centering
\includegraphics[scale=0.8]{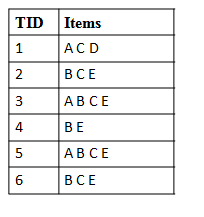}
\caption{ $Contexte D$ }
\label{figclosed}
\end{figure}

\textbf{Exemple}
La figure \ref{figclosed}  représente l'exécution de l'algorithme \textsl{Close} sur le contexte d'extraction \emph{D} pour un seuil minimal de support de $\frac{2}{6}$.

\begin{figure}[!ht]
\centering
\includegraphics[scale=0.8]{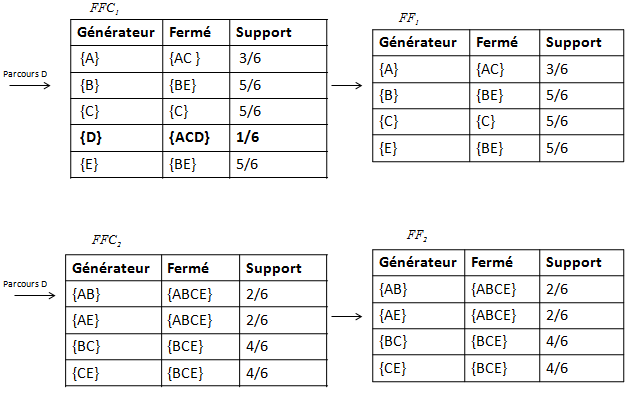}
\caption{ Extraction des itemsets fermés fréquents avec \textsl{Close} }
\label{fig4}
\end{figure}

L'ensemble $FFC_1$ est initialisé avec la liste des 1-itemsets du contexte \emph{D}. La procédure Gen-
Closure génère les fermetures des 1-générateurs, qui sont les itemsets fermés fréquents
potentiels, et leurs supports dans \begin{math}FFC_1\end{math}. Les groupes candidats de \begin{math}FFC_1\end{math} qui sont fréquents sont
insérés dans l'ensemble \begin{math}FF_1\end{math}. La première phase de la procédure Gen-Generator appliquée à
l'ensemble \begin{math}FF_1\end{math} génère six nouveaux 2-générateurs candidats : \{AB\}, \{AC\}, \{AE\}, \{BC\}, \{BE\}
et \{CE\} dans \begin{math}FFC_2\end{math}. Les 2-générateurs \{AC\} et \{BE\} sont supprimés de \begin{math}FFC_2 \end{math} par la troisième
phase de la procédure Gen-Generator car nous avons 

\{AC\} \begin{math}\subseteq \gamma \end{math} (\{A\})  et \{BE\} \begin{math}\subseteq \gamma \end{math}(\{B\}). La
procédure Gen-Closure calcule ensuite les fermetures et les supports des 2-générateurs restant
dans \begin{math}FFC_2 \end{math}et les ensembles \begin{math}FF_2 \end{math}et \begin{math}FFC_2 \end{math}sont identiques car tous les itemsets fermés de \begin{math}FFC_2\end{math}
sont fréquents. L'application de la procédure Gen-Generator à l'ensemble \begin{math}FF_2 \end{math} génère le 3-générateur \{ABE\} qui est supprimé car le 2-générateur \{BE\} n'appartient pas à \begin{math}FF_2\end{math} et l'algorithme s'arrête.

\subsubsection{Algorithme  \textsl{A-Close}}

Parmi les premiers algorithmes extrayant les  itemsets fermés fréquents nous retrouvons l'algorithme\textsl{A-Close} \cite{pasquier1999discovering}. Entre autres qualités, par rapport à Close ,après avoir construit un ensemble de k-générateurs candidats à partir des (k-1)-générateurs minimaux retenus dans la (k-1) i`eme itération. À ce niveau A-close supprime de cet ensemble tout candidat g dont le support est égal au support d'un de ses sous-ensembles de taille (k-1)

\textsl{A-Close} considère un ensemble de générateurs candidats d'une taille donnée,et détermine leurs supports et leurs fermetures en réalisant un balayage du contexte lors de
chaque itération. Les fermetures (fréquentes) des générateurs fréquents sont les itemsets fermés
fréquents extraits lors de l'itération. Les générateurs candidats sont construits en combinant les générateurs fréquents extraits durant l'itération précédente.
Ainsi, \textsl{A-Close} procède en deux étapes successives :
\begin{enumerate}

\item { il détermine tous les générateurs minimaux fréquents, c'est à dire,  les plus petits éléments incomparables des classes d'équivalence induites par l'opérateur de fermeture ã.}

\item {Pour chaque classe d'équivalence, il détermine l'élément maximal résidant au sommet de la hiérarchie. i.e l'itemset fermé fréquent.}

   \end{enumerate}




\subsubsection{ Algorithme \textsl{LCM}}
\begin{figure}[H]
\centering
\includegraphics[scale=0.8]{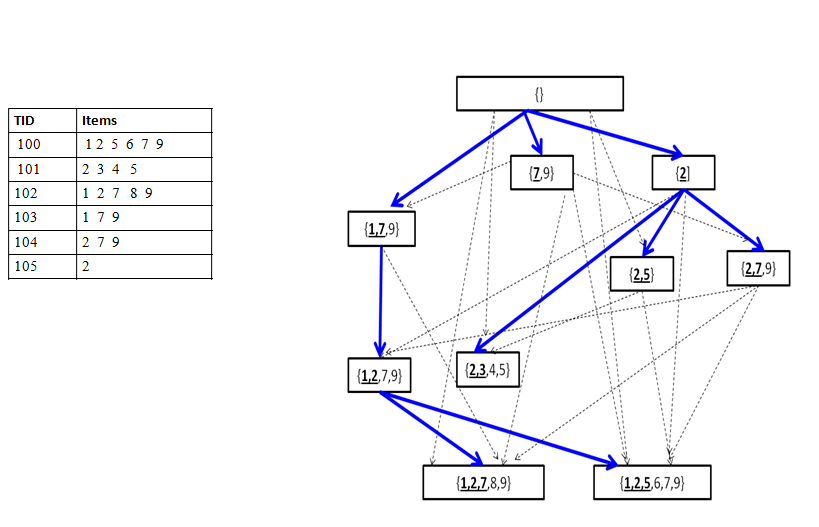}
\caption{ Exemple d’utilisation de l’algorithme LCM pour découvrir les
itemsets fréquents  }
\label{figurelcm}
\end{figure}

\textsl{LCM}  (connu sous la terminologie anglaise Linear time Closed item set Miner) a été proposé dans \cite{uno2003lcm}. Cet algorithme est consacré pour l'extraction des itemsets fermés.
\textsl{LCM} se distingue des autres algorithmes de type \textsl{backtrack}, c'est à dire, qu’il énumère linéairement l’ensemble des itemsets fréquents fermés par un parcoure en profondeur, sans explorer des motifs fréquents non nécessaires. Tel qu’illustré dans l’exemple
de la Figure \ref{figurelcm}. Pour ce faire, un arbre sous forme de trajets transversaux contenant seulement tous les motifs fermés fréquents est crée.
Deux techniques ont utilisées pour accélérer les mises à jour sur les occurrences des motifs :
\begin{enumerate}
    \item {\textbf{Occurrence deliver}: Cette technique calcule simultanément les ensembles d’occurrences de tous les successeurs du motif courant durant une, et une seule, itération de balayage sur l’ensemble d’occurrences actuel.}
\item{\textbf{Diffsets}: Cette technique a été introduite dans \cite{zaki2002charm} pour réduire l’utilisation de la mémoire des calculs intermédiaires.}
\end{enumerate}

L'algorithme \textsl{LCM} repose sur un parcours optimisé de l’espace de recherche exploitant le concept de « core prefix ».  Il faut aussi pouvoir définir un ordre sur les items (l’ordre alphabétique,par exemple). Intuitivement, le core prefix d’un itemset fermé $I$ sert de « noyau » d’extension pour générer un autre itemset fermé  \begin{math}\mathit{I^{'}} \end{math}.Le core prefix d’un itemset $I$ est le plus petit préfixe (selon l’ordre sur les items) qui apparaît dans toutes les transactions où apparaît $I$.




\textbf{Exemple}

\subsubsection{Algorithme \textsl{CLOSET}}
\begin{figure}[H]
\centering
\includegraphics[scale=0.6]{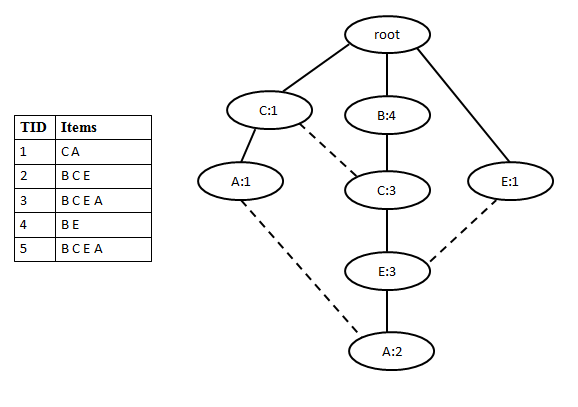}
\caption{ Contexte d'extraction avec son FP-Tree associé }
\label{figcloset}
\end{figure}

L'algorithme \textsl{CLOSET} a été proposé dans \cite{pei2000closet}. Cet algorithme utilise une structure de données avancée, basée sur la notion de trie, appelée arbre FP-Tree \cite{han2000mining}. La particularité de cette structure réside dans le fait que plusieurs transactions partagent un même chemin, de longueur n dans l'arbre FP-Tree. S'ils ont les n premiers items en commun, l'algorithme \textsl{CLOSET} effectue le processus d'extraction des itemsets fermés fréquents en deux étapes successives \cite{pei2000closet} :
\begin{enumerate}
\item{ Construction de l'arbre FP-Tree :Tel qu'illustré dans l'exemple de la Figure \ref{figcloset}. Les items des transactions sont ordonnés par support décroissant après avoir élagué les items  in-fréquents. Ensuite, l'arbre FP-Tree est construit comme suit. Premièrement, le noeud racine est créé et est étiqueté par "root". Pour chaque transaction du contexte, les items sont traités et une branche est créée suivant le besoin. Dans chaque noeud de la structure FP-Tree, il y a un compteur qui garde la trace du nombre de transactions partageant ce noeud. Dans le cas où une transaction présente un préfixe commun avec une branche dans le FP-Tree, le compteur de chaque noeud appartenant a ce préfixe est incrémenté et une sous-branche va être créée contenant le reste des items de la transaction.}
\item{Exploration de l'arbre FP-Tree : Au lieu d'une exploration en largeur d'abord des itemsets fermés candidats, \textsl{CLOSET} effectue une partition de l'espace de recherche pour effectuer ensuite une exploration en profondeur d'abord. Ainsi, il commence par considérer les 1-itemsets fréquents, triés par ordre croissant de leurs supports respectifs, et examine seulement leurs sous-contextes conditionnels (ou FP-Tree conditionnels) . Un sous-contexte conditionnel ne contient que les items qui co-occurrent avec le 1-itemset en question. Le FP-Tree conditionnel associé est construit et le processus se poursuit d'une manière récursive.}

\end{enumerate}


\subsubsection{Algorithme TITANIC}
L'algorithme TITANIC a été proposé par Stumme et al \cite{stumme2002computing} ,et qui permet de déterminer les itemsets fermés fréquents.
Le but de cet algorithme est de  minimiser le cout du calcul des fermetures Ceci est réalisé en utilisant un mécanisme de comptage par inférence \cite{bastide2000mining}.
En reposant sur la stratégie  Générer-et-tester,et qui est la mème stratégie que A-Close  .
Titanic , explore l’espace de recherche par niveau, c’est-à- dire en partant de l’ensemble vide vers les motifs de taille 1, ensuite ceux de taille 2, et ainsi de suite. 
De surcroît, cet algorithme adopte un élagage basé sur la mesure statistique minsupp.
TITANIC évite le balayage coûteux effectué par A-CLOSE pour vérifier la dernière stratégie d’élagage. Pour cela, TITANIC utilise pour chaque candidat g de taille k une variable où il stocke son support estimé, c’est-à-dire le minimum du support de ses sous-ensembles de taille (k - 1), et qui doit être différent de son support réel, sinon g n’est pas minimal. Ceci est basé sur le lemme suivant :\\
Lemme 1 :Soient X, Y \begin{math}\subseteq \end{math} I. Si X \begin{math}\subseteq \end{math} Y et $ Supp(X) = Supp(Y )$, alors \begin{math}\Lambda\end{math}(X) = \begin{math}\Lambda\end{math}(Y ).

\subsubsection{Algorithme \textsl{Prince}}
\textsl{Prince} a été proposé dans \cite{hamrouni2013efficient}, dont l’objectif principal est de pallier les principales lacunes des algorithmes dédiés à l’extraction des motifs fermés fréquents, c’est-à-dire le coût du calcul des fermetures ainsi que le fait de ne pas construire la relation d’ordre partiel\cite{hamrouni2005prince}. 
\textsl{Prince} opère en trois étapes successives :
\begin{enumerate}

    \item {Détermination des générateurs minimaux :\textsl{Prince} détermine
tous les générateurs minimaux \cite{hamrouni2010generalization} fréquents ainsi que la bordure négative non fréquente. En effet, \textsl{Prince} parcourt l’espace de recherche par niveau (et donc par taille croissante des candidats générateurs minimaux). Tout au long de ce parcours, \textsl{Prince}  élimine tout candidat g ne pouvant pas être un générateur minimal. L’élagage
d’un tel candidat est basé sur la}
   \item{Construction du treillis des générateurs minimaux.}
    \item {Extraction des bases génériques de règles\cite{bastide2000mining}.}
    
\end{enumerate}

\subsubsection{Algorithme \textsl{ZART}}
\textsl{ZART} a été proposé dans \cite{szathmary2006zart}. Un algorithme d'extraction d'itemset multifonctionnel. En effet, \textsl{ZART}  affiche un certain nombre de fonctionnalités supplémentaires et effectue les tâches suivantes, généralement indépendantes:

\begin{enumerate}
    \item {Mécanisme de comptage par inférence: cette partie de Zart est basée sur Pascal\cite{bastide2002pascal}, qui utilise les propriétés du comptage inférence \cite{bastide2000mining}. À partir d'un certain niveau, tous les générateurs peuvent être trouvés, ainsi tous les itemsets fréquents restants et leurs supports peuvent être déduits sans autre passage de base de transactions. }
    \item {Identification les itemsets fermés fréquents: cette phase consiste à identifier les itemsets fermés fréquents parmi les itemsest fréquents. Par définition : Un  motif(Itemset) fréquent  est  dit fermé  s’il ne possède aucun sur-motif qui a le même support.}
    \item{Associer les générateurs à leurs fermetures: lorsqu'un itemset fermé fréquent est trouvé, tous ses sous-ensembles fréquents sont déjà connus. Cela signifie que ses générateurs sont déjà calculés, ils doivent seulement être identifiés.}

\end{enumerate}

\subsection{Discussion}

Nous récapitulons dans le tableau \ref{tabiff} les caractéristiques des différentes approches  d'extraction séquentielle des motifs fermés fréquents étudiées.
Cette comparaison couvre les axes suivants :
\begin{enumerate}
\item{\textbf{Stratégie d'exploration:} Cette propriété décrit la stratégie d'exploration \cite{DBLP:conf/ictai/YahiaN04} des motifs générés par l'algorithme.}
   \item {\textbf{Motifs extraits :}Cette propriété décrit les motifs générés en sortie par l'algorithme}
   \item{\textbf{Caractéristiques:} Cette propriété décrit les caractéristiques de l'approche en question}
\end{enumerate}

En comparant les algorithmes décrits dans cette sous section, nous pouvons noter les remarques suivantes :

\begin{itemize}
    \item {\textbf{CLOSE, A-CLOSE et TITANIC:}ont pour désavantage de calculer la même fermeture plusieurs fois dans le cas où elle admet plusieurs générateurs minimaux.  Les stratégies d'élagage adoptées par TITANIC sont une amélioration de celle de A-CLOSE. En effet, en utilisant le support estimé d'un candidat, TITANIC évite le coUt des balayages effectués par A-CLOSE pour comparer le support d'un candidat générateur minimal de taille k aux supports de ses sous-ensembles de taille (k-1).}
    \item{\textbf{ Closet }évite le calcul dupliqué des fermetures. Ainsi il utilise  les mêmes stratégies d'élagages.}
    \item{\textbf{LCM:} LCM se distingue des autres algorithmes de type backtrack de la méthode de vérification qu’un
itemset est fermé et la méthode d’étendre un itemset fréquent fermé pour générer un nouvel itemset fréquent fermé\cite{jelassi2014efficient}. }
    \item{\textbf{Prince :} La principale originalité de PRINCE réside dans la structure de  treillis des générateurs minimaux. Ce qui permet de maintenir l'ordre partiel entre les motifs fermés fréquents  ainsi que leurs générateurs associés \cite{hamrouni2013efficient}.}
    \item {\textbf{ZART :} un algorithme d'extraction d'itemset multifonctionnel. L'idée introduite dans ZART peut être généralisée, et ainsi elle peut être appliquée à n'importe quel algorithme d'extraction d'itemset. }
    
\end{itemize}
    
    \begin{center}

    \begin{tabular}{||p{2cm}||p{2cm}||p{3cm}||p{8cm}||}
        \hline    \hline  
  Algorithme    &Stratégie d'exploration &Élément générés en sortie  & Caractéristiques\\
         \hline  \hline  
   Close\cite{pasquier1999efficient} &Générer-et-tester&Générateurs minimaux et itemsets
fermés fréquents& \begin{itemize}
    \item {Calcul  redondant de la fermeture.}
\end{itemize}.\\
    \hline\hline
    A-Close \cite{pasquier1999discovering}&Générer-et-tester&Générateurs minimaux et itemsets
fermés fréquents&\begin{itemize}
    \item {Calcul  redondant de la fermeture.}
\end{itemize}.\\
    \hline\hline
 TITANIC \cite{stumme2002computing}&Générer-et-tester&Générateurs minimaux et itemsets
fermés fréquents&\begin{itemize}
    \item {Calculer la même fermeture plusieurs fois.}
\end{itemize}.\\
     \hline \hline
    LCM \cite{uno2003lcm}&Générer-et-tester&Itemsets fermés fréquents&\begin{itemize}
    \item{type backtrack. }
    \item{Utilise la technique Diffset.}

    \end{itemize}\\
    \hline\hline    
    \end{tabular} 
 
\captionof{table}{Tableau comparatif des algorithmes d’extraction séquentielle des motifs fermés fréquents}
\label{tabifffff}
    \end{center}

    \begin{center}

    \begin{tabular}{||p{2cm}||p{2cm}||p{3cm}||p{8cm}||}
        \hline    \hline  
  Algorithme    &Stratégie d'exploration &Élément générés en sortie  & Caractéristiques\\
         \hline  \hline

    Closet \cite{pei2000closet} &Diviser pour régner&Itemsets fermés fréquents&\begin{itemize}
      \item { Mémoire centrale pour maintenir les motifs fermés fréquents.}
      \item{ FP-Tree proposée est non adaptée pour un processus de fouille interactif.}
    \end{itemize}.\\

\hline \hline
Prince \cite{hamrouni2013efficient}&Générer-et-tester&Générateurs minimaux et itemsets
fermés fréquents&\begin{itemize}
\item{Ordre partiel.}
     
    \end{itemize}.\\
    \hline\hline
ZART \cite{szathmary2006zart}& Générer-et-tester&Générateurs minimaux et itemsets
fermés fréquents& \begin{itemize}
    \item{Complet :pour le calcul des classes d'itemsets, y compris les générateurs et les itemsets fermés.}
\end{itemize}.\\
\hline\hline
    \end{tabular} 
 
\captionof{table}{Tableau comparatif des algorithmes d’extraction séquentielle des motifs fermés fréquents}
\label{tabiff}
    \end{center}

\section{Extraction parallèle des itemsets }
Malgré l’efficacité de plusieurs algorithmes séquentiels, ces algorithmes voient leurs performances se dégrader lorsque la taille des données augmente. Pour maintenir les performances de ces algorithmes, le développement d’algorithmes parallèles et distribués\cite{zitouni2017massively} apparaît comme une solution pouvant aider à accélérer la vitesse de traitement et réduire la taille d’espace mémoire utilisée. 

\subsection{Extraction des motifs fréquents en parallèle}

Dans cette sous section , nous allons passer en revue les principaux algorithmes permettant l'extraction des motifs fermés fréquents en parallèle.

 \subsubsection{Algorithme Parallel Apriori Algorithm } 
 Parallel Apriori Algorithm se base sur  l'algorithme Apriori  \cite{leskovec2014mining}.
 Dans un environnement volumineux et distribué Parallel Apriori Algorithm  la version parallèle de l'algorithme Apriori est plus performant  que
son séquentiel.\\
Même avec le paramètre de parallélisme et la disponibilité d'un nombre élevé de ressources, l'algorithme Apriori a apporté des problèmes et des limitations réguliers, comme indiqué dans sa mise en œuvre séquentielle.
Dans un environnement massivement distribué tel que MapReduce \cite{dean2008mapreduce}, utilisant l'algorithme Apriori, le nombre de jobs requis pour extraire les itemsets fréquents est proportionnel à la taille du long itemset.
Par conséquent, avec un très faible support minimum et une grande quantité de données, les performances de Parallel Apriori sont très médiocres.
Ceci est dû au fait que le processus de travail interne de l'algorithme Apriori est basé sur une approche de génération et de test candidats qui aboutit à un accès E / S disque élevé. De plus, dans un environnement massivement distribué, l'algorithme Apriori permet une communication de données élevée entre les mappeurs et les reducers, ceci est particulièrement le cas lorsque le support minimum a tendance à être très faible.
 \subsubsection{Algorithme Parallel SON  }
 
 L'algorithme SON est plus flexible et adapté pour être parallélisé dans un environnement massivement distribué.
 La version parallèle de l'algorithme SON a été proposé dans \cite{leskovec2014mining}. L'objectif de Prallel SON  est l'extraction des itemsets fréquents selon le paradigme MapReduce .
 En effet Parallel SON opère avec deux Jobs.
 \begin{itemize}
     \item{Premier Job : la base de donnée est divisées en des sous -bases,la fouilles des sous-bases s'effectue de façon parallèle à l'aide des mappers  et en utilisant un algorithme d’extraction des motifs fréquents selon une valeur minsupp locale.Ensuite les mappers ces résultats( motifs fréquents dans leurs partitions) aux reducers. Ces derniers joindraient les résultats et font la somme des valeurs de chaque clé qui sont les motifs selon l'algorithme SON puis écrivaient les résultats dans sytème de fichiers distribué de Hadoop $\emph{HDFS}$.}
     \item{Deuxième job : un classement est effectué,en séparant les motifs qui sont  globalement fréquents de ceux qui ne sont que localement fréquents.}
 \end{itemize}

 \subsubsection{Algorithme Parallel Eclat}

L'algorithme Parllel Eclat a été introduit dans \cite{zaki2000scalable}. En effet, cette version parallèle  apporte les mêmes problèmes et limitations de sa mise en œuvre séquentielle.
En particulier, le nombre élevés d'items fréquents  entraîne  une grande augmentation de nombre d'identifiant (TID) de transactions à stocker .
Cet inconvénient serait terrible  dans la capacité du mémoire, c'est à dire que la liste des identifiants de transaction ne peut pas entrer dans la mémoire disponible.

 \subsubsection{Algorithme PFP -Growth }
 PFP-Growth a été proposé dans  \cite{li2008pfp}. C'est la version   parallèle du  FP-Growth.\\
 PFP-Growth a été appliqué avec succès pour extraire efficacement les itemsets fréquents dans les grandes base de données.
 Le processus de fouille de PFP-Growth se déroule en mémoire suivant les principes suivants :
 Lors de son premier travail MapReduce, PFP-Growth effectue un processus de comptage simple pour déterminer une liste d'items fréquents.
 Le second travail MapReduce est dédié à la construction d'un arbre  FP-tree à extraire ultérieurement lors de la phase réductrice "Reduce".

\textbf{Exemple}
 Dans cet exemple une base de transactions comporte cinq transactions composées d'alphabets en minuscules et avec une valeur minsupp = 3. La première étape que FP-Growth effectue consiste à trier les items dans les transactions en supprimant les items in-fréquents.Après cette étape, par exemple, $\mathit{T1}$  (la première transaction) est élagué de $\{f, a, c, d, g, i, m, p\} à \{f, c, a, m, p\}$. FP-Growth alors
compresse ces transactions "élaguées" dans un arbre préfixe,dont lequel le racine est l'item le plus fréquent f. Chaque chemin sur l'arbre représente un ensemble de transactions qui partagent le même préfixe,chaque nœud correspond à un item. Chaque niveau de l'arborescence correspond à un item et une liste d'éléments est créée pour lier toutes les transactions qui possèdent cet item.Une fois que l'arbre a été construit, l'extraction de motifs suivante peut être effectuée.
 \begin{figure}[!ht]
\centering
\includegraphics[scale=0.8]{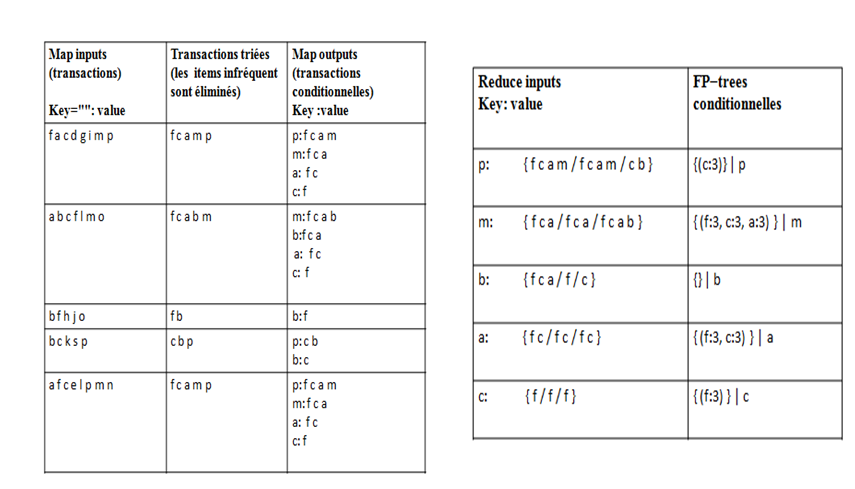}
\caption{Un exemple simple de FP-Growth distribué }
\label{figg}
\end{figure}
\\

\subsection{Extraction des motifs  fermés fréquents en parallèle}

 \subsubsection{Algorithme PLCM }
 L'algorithme PLCM a été proposé dans \cite{negrevergne2010decouverte},c'est version parallèle de l'algorithme LCM. Le but de PLCM est l'extraction des motifs fermés fréquents en parallèle. Pour ce faire, PLCM adopte un modèle où  les threads  communiquent par le biais d’un espace de mémoire partagée appelé Tuple Space auquel ils peuvent ajouter ou retirer des tuples. 
Le tuple space stoque les tuples dans N “bancs de travail”, où N est le nombre de threads utilisés. Le fait d’avoir un banc de travail assigné à chaque thread permet de limiter la contention au moment des appels aux primitives put et get. Chaque thread ajoute et consomme des tuples dans un banc qui lui est propre. Lorsque le banc d’un thread est vide, le tuple space lui donne des tuples d’un autre banc. Il s’agit d’une forme de vol de travail, qui est directement gérée par le tuple space et  transparente pour l’algorithme.

 \subsubsection{Algorithme DARCI }
L'algorithme DARCI "Distributed Association Rule mining utilizing Closed Itemsets" a été proposé dans \cite{alramouni2006darci}.Il se base sur l'algorithme CLOSET pour  l’extraction des motifs fermés fréquents locaux .Pour trouver les motifs fermés globaux ,l'algorithme DARCI implique une échange des supports locaux des itemsets fréquents localement dans chaque partitions. 
L'algorithme opère en deux phases :
 \begin{itemize}
    \item {Phase 1 :Envoi des itemsets fermés fréquents.}
     \item {Phase 2 :DARCI utilise une technique  qui s’appel ‘best scenario pruning’ pour envoyer les supports locaux aux autres partitions ’, si un itemset(motif)
I est localement fréquent dans une partition $P_i $alors il est diffusé dans le cadre des itemsets fréquents locaux, mais si I est non fréquent dans la partition $P_i$, $ P_i$  doit décider la diffusion du support local de l'itemset I ou non. $P_i$
 ne devrait pas diffuser le support de I si ce dernier n’est pas fréquent globalement dans le meilleur scénario possible.}
\end{itemize}

$P_i$ diffuse le support de I si $Sup_best (I) \geq  \sigma$ tel que $\sigma$ est le support minimum global,et le $Sup_best$  est le  meilleur scénario possible du support global de I.
 \begin{center}
     
     $$  Sup(\propto) \leq  sup_best(I) \ Tel \ que $$\\
     $$    sup_best(I)=\sum I \ freq \  Supp^{i}(I) + \sum I \ nonfreq \ (\ Minsupp*\left | P^{i} \right |-1)$$

 \end{center}

\subsection{Discussion}
Le tableau \ref{tabsp} présente une comparaison entre les différents algorithmes que nous avons
présenté ci-dessus. La comparaison est faite selon les axes suivants :

\begin{itemize}
    \item {\textbf{Algorithme de base }}
    \item{\textbf{Type de motifs extraits }}
        \item {\textbf{Version parallèle }}
\end{itemize}

 \begin{center}

    \begin{tabular}{||p{4cm}||p{8cm}||p{4cm}||}
        \hline \hline      
Algorithme de base  &   Types de motifs     & Parallèle  \\
         \hline  \hline   
Apriori & motifs fréquents & Parallel  Apriori \\  
\hline \hline 
FP-Growth & motifs fréquents &PFP-Growth\\
\hline \hline 
Eclat &  motifs fréquents&Parallel Eclat\\
\hline \hline 
SON & motifs fréquents  & Parallel SON \\
\hline \hline 
Close & fermés fréquents +générateurs minimaux &  - \\
\hline \hline 
A-Close & fermés fréquents +générateurs minimaux & -\\
\hline \hline 
Closet & motifs fermés fréquents & DARCI\\
\hline \hline 
TITANIC & fermés fréquents +générateurs minimaux & -\\
\hline \hline 
ZART & fermés fréquents +générateurs minimaux & -\\
\hline \hline 
LCM& motifs fermés fréquents & PLCM\\
\hline \hline 
\end{tabular} 
 
\captionof{table}{Tableau comparatif des algorithmes d’extraction des motifs}
\label{tabsp}
    \end{center}

\section{Classification des algorithmes distribués d’extraction des motifs fréquents et des motifs  fermés fréquents }

Un premier examen de ces algorithmes permet de les classer selon la stratégie de partitionnement, à savoir le partitionnement des données et le partitionnement de l’espace de recherche. 
Nous présentons dans cette section ces caractéristiques permettant de montrer les différences majeures qui pourraient exister entre les algorithmes parallèles et distribués passés en revue.
\begin{itemize}

\item{\textbf{ Stratégie de partitionnement }: Deux stratégies de partitionnements ont été avancées : un partitionnement des données et un partitionnement de l’espace de recherche.}

\item{\textbf{ Technique d’exploration }: Il existe deux techniques d’exploration de l’espace de recherche, à savoir « tester et générer » et « diviser pour régner ».}
\item{\textbf{ Algorithme de base} : Se sont des algorithmes séquentiels utilisés pour extraire les itemsets fréquents locaux et les itemsets fermés fréquents locaux.}
\item{\textbf{ Type de motifs extraits }: Se sont les motifs générés en sortie}

\end{itemize}

\begin{center}

    \begin{tabular}{|p{3cm}|p{2cm}|p{2cm}|p{2cm}|p{2cm}|p{2cm}|p{2cm}|}
        \hline      
Algorithmes&PLCM  &  Parallel Apriori  & Parallel SON& Parallel Eclat &PFP-Growth& DARCI\\
         \hline    
Stratégie de partitionnement & Espace de recherche&Espace de recherche
 &Espace de recherche&Espace de recherche&Données horizontal& Données horizontal\\  
\hline
Technique d’exploration & Diviser pour régner & Tester et générer &Tester et générer&Diviser pour régner&Diviser pour régner&Diviser pour régne. 
\\
\hline 
Algorithme de base &LCM & Apriori &SON&Eclat&FP-Growth&Closet \\
\hline
Type de motifs extraits & motifs fermés fréquents &motifs fréquents &motifs fréquents&motifs fréquents&motifs fréquents &motifs fermés fréquents.\\

\hline
\end{tabular} 
 
\captionof{table}{ Classification des algorithmes  d'extraction parallèles des motifs fermés fréquents}
\label{ta}
    \end{center}

\section{Conclusion}
\hspace{3ex} Dans ce chapitre nous avons discuté les problèmes reliés au processus d’extractions des motifs fréquents et des motifs fermés fréquents. En étudiant les approches proposées dans l’état de l’art les avantages et les limitations de ces processus d’extraction des motifs sont reliés particulièrement aux accès multiples à la base des données et la capacité de la mémoire.
Typiquement, ces différents limitations présentent un défi majeur quand le volume des données est énorme et que le support minimum est très petit ou que les motifs à découvrir sont de grande taille.

À cet égard, nous proposons une nouvelle  approche permettant l'extraction  des motifs fermés fréquents avec leurs générateurs minimaux associés.

\newpage

\chapter{Une nouvelle approche pour l'extraction des itemsets fermés fréquents   }
\section{Introduction}

Pour pallier aux insuffisances des algorithmes séquentiels, la recherche  simultanée d’itemsets fermés fréquents en partitionnant l'espace de recherche apparaît comme une solution intéressante.




Dans ce chapitre, nous allons introduire  dans la première section  le principe de l'approche proposée. Dans la deuxième section, nous allons présenter l'architecture globale de notre approche. La troisième section est dédiée à la conception  détaillée de notre approche. Enfin la quatrième section est consacrée à présenter un exemple illustratif de  notre approche.
\section{Principe de l'approche}
Dans cette section, nous introduisons le paradigme "diviser pour régner", c'est  principe sur lequel se base notre approche.

\subsection{présentation du paradigme "Diviser pour régner"}
En informatique, diviser pour régner ( "divide and conquer" en anglais) est une technique algorithmique consistant à :
\begin{itemize}
    \item {\textbf{Diviser} : découper un problème initial en sous-problèmes}
    \item{\textbf{Régner} : résoudre les sous-problèmes.}
    \item{\textbf{Combiner} : calculer une solution au problème initial à partir des solutions des sous-problèmes.}
\end{itemize}
La méthode de diviser pour régner est une méthode qui permet, parfois de trouver des solutions efficaces à des problèmes algorithmiques. L’idée est de découper le problème initial, de taille n, en plusieurs sous-problèmes de taille sensiblement inférieure, puis de recombiner les solutions partielles.\\
De façon informelle, il s'agit de résoudre un problème de taille n à partir de la résolution de deux instances indépendantes du même problème mais pour une taille inférieure. De nombreux problème peuvent être résolus de cette façon. À cet égard, nous allons diviser la base des transactions $\emph{D}$, puis nous allons appliquer sur chaque partition(sous-base ou sous-contexte) de la base des transactions  un algorithme d'extraction des motifs fermés fréquents ainsi leurs générateurs minimaux  pour avoir un fichier pour chaque partition contenant  des motifs fermés fréquents avec les générateurs minimaux. Ensuite deux algorithmes que nous avons proposé UFCIGs et UFCIGs-pruning seront appliqué  pour combiner les fichiers par deux. En effet, ces algorithmes permettent la mise à jours des motifs fermés fréquents avec leurs générateurs minimaux .

\section{Conception globale  de l'approche UFCIGs-DAC }
Les algorithmes de ce type apparaissent comme composés de deux algorithmes; le premier partage le problème en sous-problèmes, le second algorithme fusionne les résultats partiels en  résultat global. 
Donc l'architecture de notre approche comporte deux grandes phases .
\begin{enumerate}
    \item {\textbf{Phase 1} : Phase de partitionnement de la base et d'extractions des Itemsets fermés fréquents avec leurs générateurs minimaux.}

    \item{\textbf{Phase 2 }: c'est la phase du fusionnement des résultats partiels; la mise à jour des motifs fermés fréquents et des  générateurs minimaux   }
    \end{enumerate}
Le graphe résultant de cette architecture est comme suit:

 \begin{figure}[H]
\centering
\includegraphics[scale=0.8]{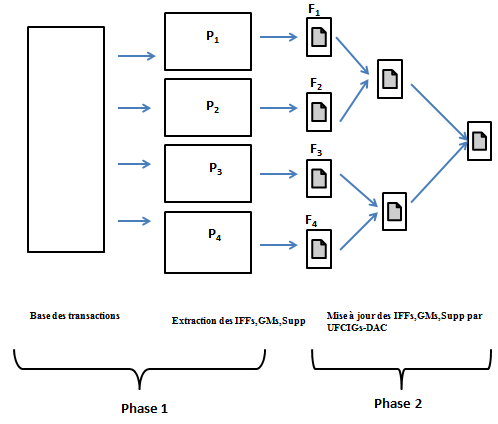}
\caption{Conception globale de l'approche UFCIGs-DAC   }
\label{fig44}
\end{figure}
\section{Conception détaillée de l'approche}
Dans cette section, nous allons détailler les deux phases que nous les avons énoncé dans la section précédente.

\subsection{Phase1}
Cette phase est destinée à préparer les entrées de l'algorithme UFCIGs-DAC de la deuxième phase. D'abord, un partionnement de la base de transactions $\emph{D}  \ en \left | P \right |= n $ partitions de transactions (par exemple n=4).
$\emph{D} =\{P_1,P_2,P_3,P_4\}$. Ensuite, le processus d'extraction des Itemsets fermés fréquents avec leurs générateurs minimaux  sera exécuté sur chaque partition. Enfin, les fichiers $ F_1 ,F_2,F_3, \ et \ F_4 $ sont respectivement les résultats d'extractions des partitions $P_1,P_2,P_3, \ et\ P_4$ .\\


\subsection{Phase 2}
Après avoir récupéré les fichiers \begin{math} F_1 ,F_2,F3, et F_4 \end{math}, l'algorithme \textsl{UFCIGs} sera appliqué pour faire la mise à jour des Itemsets fermés fréquents avec leurs générateurs minimaux. Ce processus sera effectué entre chaque deux fichiers pour fournir en sortie un seul fichier résultat global.

\subsubsection{Présentation de l'algorithme UFCGs-DAC}

Réduire l'espace de recherche  (le nombre des transactions )en partitionnant la base des transactions totale  en des sous-bases. Sur chaque sous-base  nous  appliquons un algorithme d'extractions des Itemset fermés fréquents et des générateurs minimaux, puis récolter les résultats en un seul sans prendre en considérations les propriétés de la fermeture et de générateur minimal ça va donner des résultats erronés certainement.\\
UFCIGs-DAC "Update of frequent closed itemsets and their generators" est un algorithme de mise à jour des itemsets fermés fréquents et  des générateurs minimaux  selon la stratégie "Divide And Conquer" .
L'objectif principal de \textbf{ UFCIGs-DAC} est de recombiner chaque deux  solutions partielles toute en respectant les notions de fermeture, de générateur minimal avec leur valeur de support  que nous avons déjà présenté dans le premier chapitre "notions de base".
nous rappelons les propriétés suivantes : 
\begin{itemize}
     
      \item{\textbf{Propriété 1}: un motif $I \subseteq  \emph{I}$ est dit fréquent si son support relatif, Supp($\textit{I}$)=$\frac{\left | \psi (I) \right |}{\left | \mathrm{O} \right |}
 $  dépasse un seuil minimum fixé par l’utilisateur noté minsupp.Notons que $\left | \psi (I) \right |$ est appelé support absolu de ($I$)\cite{hamrouni2011construction}. }

\item{\textbf{Propriété 2} : Un itemset (motif)  $I \subseteq I$ est dit fermé si $ I =\lambda(I)$. L'itemset $\mathit{I}$ est un ensemble maximal  l'items communs un ensemble d'objets \cite{hamrouni2011construction}.}

 \item{ \textbf{Propriété 3 }:  Un itemset g  \begin{math} \subseteq \end{math}$\mathfrak{I}$ est un générateur  minimal d’un itemset fermé  $\textit{I}$ si  et seulement si  $\lambda(g) = \textit{I}$et \begin{math} \nexists \end{math} g' \begin{math} \subseteq  \end{math} g  tel  que $ \lambda(g') = \textit{I}$ \cite{hamrouni2011construction}.}

\end{itemize}

\subsubsection{Description de l'algorithme principal de l'approche UFCIGs-DAC }
Les notations utilisées sont résumées dans le tableau \ref{tabnotation}.

 \begin{center}

    \begin{tabular}{|p{4cm}|p{12cm}|}
        \hline          \hline  
  IFFs   &   Itemsets (motifs) fermés fréquents\\
         \hline        \hline  
GMs      &   Générateurs minimaux.\\  
\hline      \hline  
 Supp &  Le support d'un motif.\\
\hline     \hline  
$CL_1$ &     Liste de closed ,support et générateurs minimaux du fichier $F_1$ .\\
\hline     \hline  
$CL_2$ &     Liste de closed ,support et générateurs minimaux du fichier $F_2$ .\\

\hline\hline 
$CL_1.Listclosed$& Liste de closed du fichier $F_1$.\\
\hline     \hline  
$CL_1.Listgm$& Liste de générateurs minimaux du fichier $F_1$.\\

\hline     \hline  

$CL_2.Listclosed$& Liste de closed du fichier $F_2$.\\
\hline     \hline  
$CL_2.Listgm$& Liste de générateurs minimaux du fichier $F_2$.\\

\hline     \hline  

Supp-Abs &  Support absolu.\\
\hline      \hline  
Supp-BitSet-Rpr &  Support calculé avec une représentation en Bitset.\\
\hline     \hline  
LFres  & Liste de  closed, support et générateurs minimaux du résulat final . \\
\hline     \hline

\end{tabular} 
 
\captionof{table}{les notations adoptées dans l'algorithme UFCIGs-DAC.}
\label{tabnotation}
    \end{center}

\begin{figure} [H]
\centering
\includegraphics[scale=1.0]{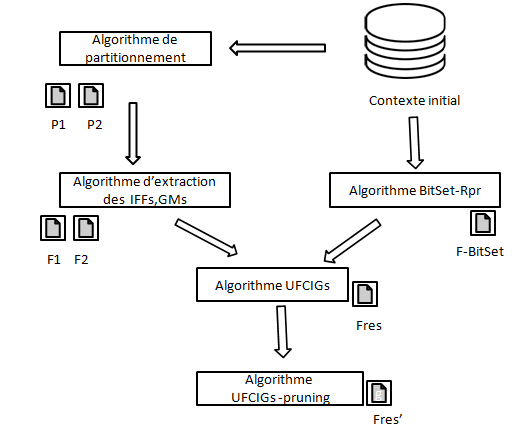}
\caption{L'algorithme principal de l'approche UFCIGs-DAC }
\label{figmain}
\end{figure}

L'algorithme principal de notre approche est présenté dans l'algorithme \ref{mainalgo} et la Figure \ref{figmain}. Ces derniers décrivent le processus de déroulement de notre approche, c'est à dire les appels entre les  différents algorithmes de notre approche .

\begin{algorithm}[H]
\label{mainalgo}
\small
\CommentSty{Begin}\\
\CommentSty{Algorithme Partitionnement ($Contexte\_init$) ;} \\
\CommentSty{Algorithme \textbf{Bitset-Rpr}($Contexte\_init$);}\\
\CommentSty{Algorithme d'extraction des IFFs et GMs($P_1, P_2$);}\\
\CommentSty{Algorithme \textbf{UFCIGs}($F_1, F_2$);}\\
\CommentSty{Algorithme \textbf{$UFCIGs\_pruning$}(Fres) ;}\\
\CommentSty{End.}
\caption{Main UFCIGs-DAC}
\end{algorithm}

\newpage
\subsubsection{Description détaillée des différentes algorithmes de l'approche UFCIGs-DAC}
Dans cette sous-section, nous allons décrire de façon détaillée et ordonnée les appels entre les différents composants "algorithmes" de notre approche.

\begin{enumerate}
\item{\textbf{Algorithme Partitionnement}}\\
L'algorithme Partitionnement permet de partitionner le contexte initial \textsc{$Contexte\_init$} 
et prend en entrée :
\begin{itemize}
    \item {\textsc{$Context\_init$ }: représentant le fichier du contexte initiale, c'est à dire la base des transactions totale.}
   
   \end{itemize}
Pour obtenir les  partitions: 
\begin{itemize}
    \item {$P_1$: la première partition ou (sous-contexte).}
    \item  {$P_2$: la deuxième partition ou  (sous-contexte).}
\end{itemize}

\item{\textbf{Algorithme  BitSet-Rpr }}\\

En se basant sur l'algorithme \textsl{$AprioriTID\_BitSet$} \cite{agrawal1994fast}, qui permet de calculer les  itemsets fréquents dans une base de transactions. Cet algorithme utilise les bitsets comme structures internes pour représenter les TIDs des transactions. En effet, l'avantage de  l'algorithme \textsl{$AprioriTID\_BitSet$} réside dans  l'utilisation de BitSets permettant de représenter les ensembles d'identifiants de transactions de façon efficace en termes de consommation de mémoire permettant  l'exécution de l'intersection de deux ensembles d'identifiants de transactions (TIDs)  efficacement avec des ensembles de bits. Nous allons proposer un algorithme \textsl{BitSet-Rpr} qui se base sur l'algorithme \textsl{$AprioriTID\_BitSet$}, toutefois  nous nous intéressons à la représentation des TIDs des items en BitSets.\\

 \textsl{$BitSet-Rpr$} prend en entrée :
\begin{itemize}
    \item { \textsc{$Context\_init$ }: représente contexte initiale, c'est à dire la base des transactions totale.}
\end{itemize}
Pour retourner le fichier:
\begin{itemize}
    \item {\textsl{F\_Bitset:} correspond au fichier contenant tous les items du contexte initial, ainsi que leurs TIDs.}
\end{itemize}

Cela permet de calculer les supports Supp-Biset-Rpr et Sup-Abs par \textbf{l'algorithme 2}.

\item{\textbf{ Algorithme d'extraction des IFFs,GMs}}\\
Aprés avoir partitionné le contexte intial, nous allons appliquer un algorithme d'extraction spécifique (par exemple ZART \cite{szathmary2006zart}) sur chaque  partition $P_1$ et $P_2$, pour obtenir les fichiers $F_1$ et $F_2$.
Chacun de ces fichiers contient les iemsets fermés fréquents, ainsi que leurs générateurs minimaux associés.

\item{\textbf{Algorithme UFCIGs}}\\
Une fois que le processus d'extraction des partions est terminé,
l'algorithme \textsl{UFCIGs}, commence à faire la mise à jour de l'ensemble des itemsets fermés fréquents et des générateurs minimaux.
L'algorithme \textsl{UFCIGs}, dont le pseudo-code est décrit par les algorithmes 2 et 3 prend  en entrée :
\begin{itemize}
    \item {$CL_1$: Liste des objets de type \textsc{ClosedSuppGen}, c'est à dire, l'objet \textsc{ClosedSuppGen} qui contient les attributs Closed, Support, et générateur du fichier $F_1$.}
    \item{$CL_2$:  Liste des objets de type  \textsc{ClosedSuppGen}. c'est à dire, l'objet \textsc{ClosedSuppGen} qui contient les attributs Closed, Support, et générateur du fichier $F_2$.}
\textsl{UFCIGs} commence à parcourir les deux listes $CL_1$ et $CL_2$ comme suit:
\begin{itemize}
    \item {Si un IFF (Closed $c_1$) de la liste des motifs fermés  fréquents   \textsc{Listclosed} de $LC_1$ apparaît dans  la liste des motifs fermés  fréquents \textsc{Listclosed} ou la liste des générateurs minimaux \textsc{Listgm} de $LC_2$, dans ce cas \textsl{UFCIGs}, stocke $c_1$ dans la liste LFres avec une valeur de support égale à la somme des deux supports.}
    \item{Si un IFF (Closed $c_1$) de la liste des motifs fermés  fréquents   \textsc{Listclosed} de $LC_1$ n'apprait pas ni dans  la liste des motifs fermés  fréquents \textsc{Listclosed} ni dans la liste des générateurs minimaux \textsc{Listgm} de $LC_2$, dans ce cas \textsl{UFCIGs} faire le calcule de support( $Supp\_BitSet\_Rpr (c_1)$ ) en Bitset, c'est à dire en faisant l'intersection des TIDs qui sont représentés dans le fichier $F\_BitSet$. Aprés avoir calculé le Support  \textsl{$Supp\_BitSet\_Rpr(c_1)$} en Bitsets, \textsl{UFCIGs} vérifie si son support $Supp\_BitSet\_Rpr$ est supérieur ou égale au support absolu du contexte initial. Ceci est expliqué par le fait qu'un itemset fermé fréquent localement se trouvant dans une seule partition peu ne plus l'être globalement. Si le  $Supp\_BitSet_Rpr(c_1) \geq  Supp\_Abs$, \textsl{UFCIGs} stocke $c_1$ dans la liste LFres avec une la valeur de support calculée en BitSets, sinon il ne l'enregistre pas dans LFres.}
    \item{Le même processus est appliqué avec les IFFs de $CL_2$.}
    \item {Aprés avoir visité tous les IFFs des Listclosed de $CL_1$ et $CL_2$, \textsl{UFCIGs} se déclenche à traiter les GMs ($gm_1$) de la liste des générateurs minimaux \textsc{Listgm} de $LC_1$ : Si un $gm_1$ de $LC_1$ n'apparaît pas ni dans la Listclosed de $LC_1$ et $LC_2$, c'est à dire qu'il n'est pas traité : Si $gm_1$ de la \textsc{Listgm} de $LC_1$ apparaît dans la  \textsc{Listgm} de $LC_1$, stocke $gm_1$ dans la liste LFres en tant qu'un itemset fermés fréquent avec une valeur de support égale à la somme des deux supports. Sinon, \textsl{UFCIGs} calcule son support en BitSets, si $Supp\_BitSet\_Rpr(gm_1) \geq  Supp\_Abs$, dans ce cas, $gm_1$ sera stocké dans LFres en tant qu'un itemset fermés fréquent avec un valeur de support calculé en BitSets, sinon il ne l'enregistre pas dans LFres.  }
      \item{Le même processus est appliqué avec les GMs de $CL_2$.}
\end{itemize}
L'algorithme \textsl{UFCIGs} génère en sortie :
\item {LFres: Liste des itemsets  qui sont globalement fermés fréquents .}
\end{itemize}

\item{\textbf{ Algorithme  UFCIGs pruning  :}}\\
Après avoir récupéré le fichier Fres par \textbf{l'algorithme UFCIGs}, une étape d'élagage doit être appliquée pour répondre à la propriété de fermeture, puis générer et affecter les générateurs minimaux aux itemsets fermés fréquents correspondants. Ceci est effectué par l'algorithme  \textbf{ UFCIGs pruning} dont le pseudo-code est décrit par l'\textbf{algorithme 4}  qui prend en entrée:
\begin{itemize}
    \item {\textbf{LFres:} Liste Fres de \textbf{l'algorithme UFCIGs } contenant l'ensemble des itemsets fermés fréquents après leurs mise à jour. }
\end{itemize}
Pour donner en sortie :
\begin{itemize}
    \item {LFres: Liste finale  contenant l'ensemble des itemsets fermés fréquents ainsi que leurs générateurs minimaux après leurs mises à jour.}
\end{itemize}

L'algorithme commence par :
\begin{enumerate}
    \item {Tester la notion de fermeture comme suit: si un $ IFF \subset  dans IFF^{''} et supp(IFF)= supp(IFF^{'})$, alors IFF ne sera plus considéré comme un motif fréquent fermé selon la \textbf{propriété 2}, donc il sera supprimé de la lise LFres.}
    \item{affecter les GMs aux IFFs restants dans LFres à travers la procédure $\emph{Find generators}$. Cette procédure associe les  items ou les itemsets aux IFFs  adéquats comme des GMs s'ils vérifient  la \textbf{propriété 3} comme suit: \\
    Si un item $I  \subset  IFF$  et $ supp(I)= supp(IFF)$ et $I =I-plus-petit$ ( itemset composé de  moins d'items). Sinon, le motif fermé fréquent confond avec son générateur minimal, c'est à dire $IFF=GM$.}
\end{enumerate}
L'algorithme s'arrête lorsque il n'y a plus des IFFs à visiter.

\end{enumerate}

\newpage

\section{Exemple illustratif}
Dans cette section, nous allons présenter un exemple illustratif  qui décrit les différentes pahses de notre approche. 
Considérons la base des transactions \emph{D} décrit par la Figure \ref{fig4}, avec le choix de la valeur de support relatif  comme mesure de fréquence est minsupp =0.6. 

 \begin{figure}[H]
\centering
\includegraphics[scale=1.3]{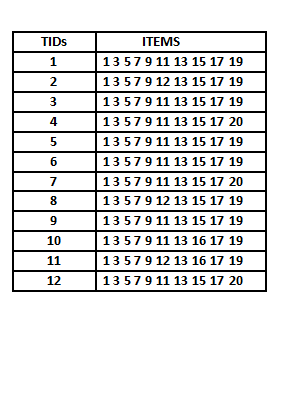}
\caption{Base des transactions \emph{D}}
\label{fig4444}
\end{figure}
\begin{figure}[H]
\centering
\includegraphics[scale=0.8]{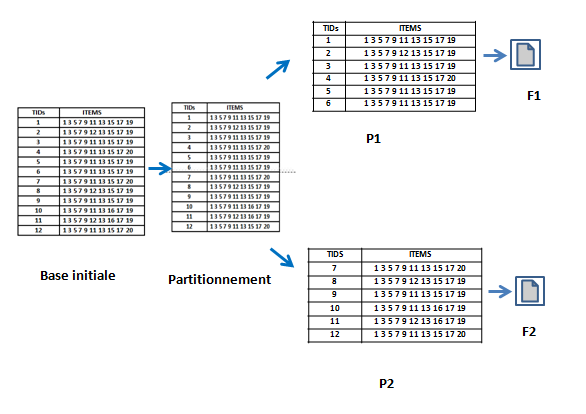}
\caption{Exemple illustratif de la conception globale de l'approche UFIGs-DAC }
\label{figproc}
\end{figure}

Tout d'abord, nous allons  partitionner la base des transactions $\emph{D}$, par exemple en deux partitions  $P_1$ et $P_2$, puis nous allons  appliquer sur chaque partition un algorithme spécifique à l'extraction des IFFs avec leurs GMs associés, pour avoir les fichiers $F_1$ et $F_2$.\\
Le processus est est représenté  dans la Figure \ref{figproc} ci-après.

La Figure \ref{figp1p2} montre les deux partitions  $P_1$ et $P_2$. Chaque partition contient six transactions. En fixant une valeur de minsupp=0.6. A cet étape nous allons appliquer un algorithme (par exemple ZART\cite{szathmary2006zart}) pour extraire les IFFs, GMs et les valeurs de support sur les partitions  $P_1$ et $P_2$.
 \begin{figure}[H]
\centering
\includegraphics[scale=0.8]{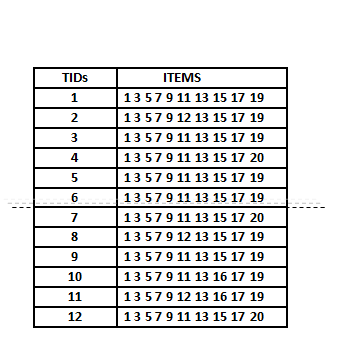}
\caption{Partitionnement de la  base des transactions \emph{D} }
\label{fig44444}
\end{figure}

\begin{figure}[H]
\centering
\includegraphics[scale=0.8]{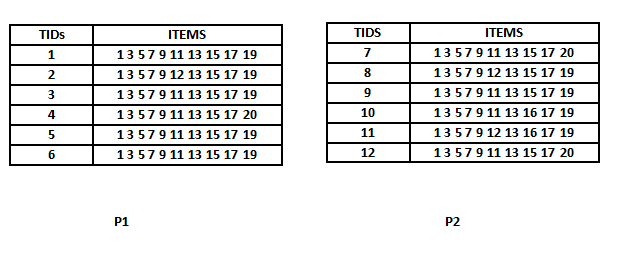}
\caption{ Les deux partitons  $P_1$ et $P_2$ }
\label{figp1p2}
\end{figure}

\begin{figure}[H]
\centering
\includegraphics[scale=0.9]{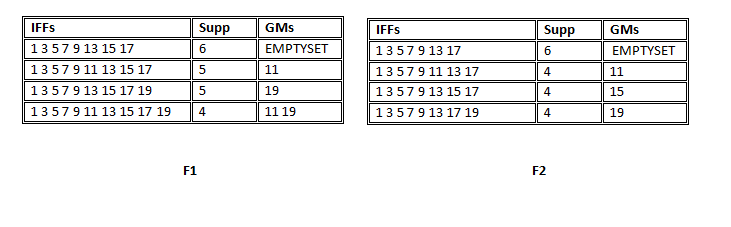}
\caption{ Les fichiers $F_1$ et $F_2$ }
\label{figf1f2}
\end{figure}

Les résultats de la première phase sont représentés dans la Figure \ref{figf1f2}. Le fichier  $F_1$ est le résultat de la partition $P_1$  contenant l'ensemble des IFFs avec leur valeur de support et leurs GMs, alors que le fichier $F_2$ est le résultat de la partition  $P_2$ contenant l'ensemble des IFFs avec leur valeur de support et leurs GMs.
 
 Le tableau \ref{tabbit} est le résultat de  l'algorithme \textsl{BitSet\_Rpr}. c'est la représentation en BitSet de tous les items de la base des transaction \emph{D}.
L'idée de cet algorithme est de déterminer  les "TIDs" de chaque item . En outre ,BitSet-Rpr détérmine les transactions dont lesquelles est apparu un item i.
Par exemple  l'item "1" est apparu dans les transactions $\{1,2,3,4,5,6,7,8,9,10,11,12 \}$, l'item "16" est apparu dans les transactions $\{9,10\}$.

 \begin{center}

    \begin{tabular}{||p{2cm}|p{6cm}||}
        \hline   \hline   
  ITEMS    &   TIDS\\
         \hline     \hline  
1 & 1 2 3 4 5 6 7 8 9 10 11 12\\  
\hline  \hline  
3 &  1 2 3 4 5 6 7 8 9 10 11 12\\
\hline   \hline  
5 & 1 2 3 4 5 6 7 8 9 10 11 12\\
\hline   \hline  
7 & 1 2 3 4 5 6 7 8 9 10 11 12\\
\hline  \hline  
9 &  1 2 3 4 5 6 7 8 9 10 11 12 \\
\hline  \hline  
11 & 1 3 4 5 6 7 9 10 12\\
\hline   \hline  
12 & 2 8 11 \\
\hline  \hline  
13 & 1 2 3 4 5 6 7 8 9 10 11  12   \\
\hline  \hline  
15 &  1 2 3 4 5 6 7 8 9 12           \\
\hline  \hline  
16 &  9 10     \\
\hline  \hline  
17 &  1 2 3 4 5 6 7 8 9 10 11  12   \\
\hline  \hline  
19 & 1 2 3 5 6 8 9 10 11 \\
\hline   \hline  
20 & 4 7 12 \\
\hline  \hline  

\end{tabular} 
 
\captionof{table}{Fichier  $F-BitSet$ }
\label{tabbit}
    \end{center}

L'algorithme \textsl{ UFCIGs} prend en entrée les fichiers $F_1 ,F_2$ et $F\_BitSet$ .

UFCIGS commmence à parcourrir les IFFs des $F_1$ et $F_2$ (cf.Figure \ref{figf1f2}).

\begin{itemize}
    \item \textcolor{red}{"1 3 5 7 9 13 15 17"} appartient à la liste des IFFs de $F_2$, alors il sera inséré dans le fichier Fres avec une valeur de support égale à la somme des deux supports, c'est à dure (6+4 = 10). Les itemsets (IFFs ou GMs) de ce type sont colorés en rouge.
    \item \textcolor[rgb]{0,0.5,0}{"1 3 5 7 9 11 13 15 17"} n'appartient ni à la liste des IFFs , ni à la liste des GMs de $F_2$. Dans ce cas, son support sera calculé en BitSet.\\
   $ Supp\_BitSet\_Rpr(\textcolor[rgb]{0,0.5,0}{\ 1 \ 3 \ 5 \ 7 \ 9 \ 11 \ 13 \ 15 \ 17})=8 \geq Supp\_Abs =8$. Alors,  \textcolor[rgb]{0,0.5,0}{"1 3 5 7 9 11 13 15 17"} sera inséré dans Fres avec une valeur de support =8.
   \item \textcolor[rgb]{0,0.5,0}{" 1 3 5 7 9 13 15 17 19"} n'appartient ni à la liste des IFFs , ni à la liste des GMs de $F_2$. Dans ce cas, son support sera calculé en BitSet.\\
       $Supp\_BitSet\_Rpr(\textcolor[rgb]{0,0.5,0}{\ 1 \ 3 \ 5 \ 7 \ 9  \ 13 \ 15 \ 17 \ 19})=6 <Supp\_Abs$. Alors,  \textcolor[rgb]{0,0.5,0} {" 1 3 5 7 9 13 15 17 19"} ne sera pas inséré dans Fres. De même pour \textcolor[rgb]{0,0.5,0}{"1 3 5 7 9 11 13 15 17 19"}.
       \item \textcolor{blue}{"1 3 5 7 9 13 17"} n'appartient ni à la liste des IFFs , ni à la liste des GMs de $F_1$. Dans ce cas, son support sera calculé en BitSet.\\
     $ Supp\_BitSet\_Rpr(\textcolor{blue}{\ 1 \ 3 \ 5 \ 7 \ 9 \ 13 \ 15 \ 17 \ 19})=12 \geq Supp\_Abs =8$. Alors,  \textcolor{blue}{"1 3 5 7 9 13 15 17 19"} sera inséré dans Fres avec une valeur de support =12. De même pour les IFFs \textcolor{blue}{"1 3 5 7 9 11 13 17"} et \textcolor{blue}{"1 3 5 7 9 13 17 19"}.
\end{itemize}

 \begin{figure}[H]
\centering
\includegraphics[scale=0.8]{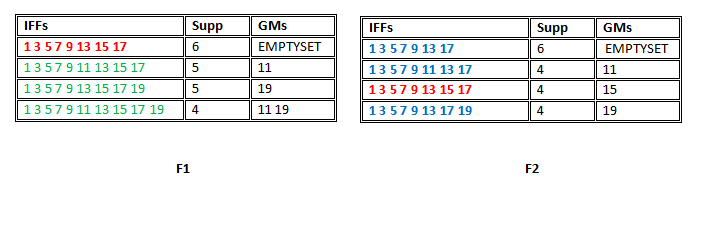}
\caption{ Parcoure des IFFs des fichiers $F_1$ et $F_2$}
\label{figf1f2clr}
\end{figure}

 \begin{figure}[H]
\centering
\includegraphics[scale=0.8]{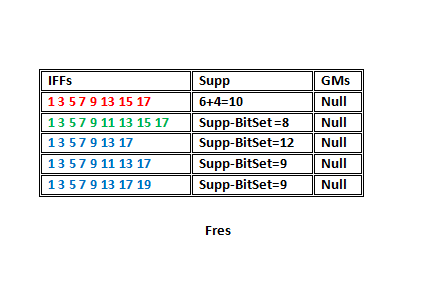}
\caption{Fichier résultat \textsl{Fres}}
\label{fresbefore}
\end{figure}

Une fois, que tous les IFFs de $F_1$ et de $F_2$ sont traités,nous allons passer au traitement des GMs de  $F_1$ et de $F_2$ .

 \begin{figure}[H]
\centering
\includegraphics[scale=0.8]{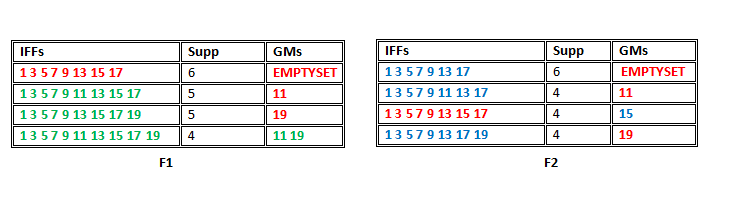}
\caption{ Parcoure des GMs des  fichiers $F_1$ et $F_2$}
\label{fres}
\end{figure}

\begin{itemize}
    \item \textcolor{red}{EMPTYSET} appartient à la liste des  GMs de $F_2$, alors il sera inséré dans le fichier Fres avec une valeur de support égale à la somme des deux supports, c'est à dure (6+6 = 12). 
    \item \textcolor{red}{11} appartient à la liste des  GMs de $F_2$, alors il sera  inséré dans le fichier Fres avec une valeur de support égale à la somme des deux supports, c'est à dure (5+4 = 9). 
    \item \textcolor{red}{19} appartient à la liste des  GMs de $F_2$, alors il sera insérér dans le fichier Fres avec une valeur de support égale à la somme des deux supports, c'est à dure (5+4 = 9). 
    \item \textcolor [rgb]{0,0.5,0}{11 19}  n'appartient  à la liste des IFFs des GMs de $F_2$. Dans ce cas, son support sera calculé en BitSet.\\
       $Supp\_BitSet\_Rpr(\textcolor [rgb]{0,0.5,0} { \ 11 \ 19})=6 <Supp\_Abs$. Alors,  \textcolor[rgb]{0,0.5,0} {" 11 19"} ne sera pas inséré dans Fres. 
       \item \textcolor{blue}{"15"} n'appartient  à la liste des IFFs des GMs de $F_1$. Dans ce cas, son support sera calculé en BitSet.\\
       $Supp\_BitSet\_Rpr(\textcolor{blue}{15})=10 >=Supp\_Abs$. Alors,  \textcolor{blue} {"15"} sera  inséré dans Fres. 
\end{itemize}

 \begin{figure}[H]
\centering
\includegraphics[scale=0.8]{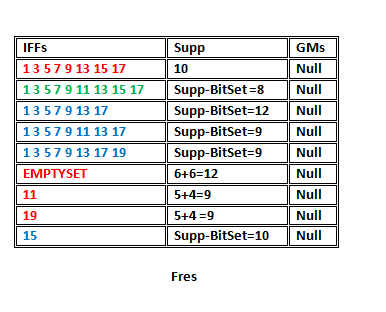}
\caption{ Fichier Fres de l'algorithme UFCIGs}
\label{fresbf}
\end{figure}

Aprés avoir traiter tous les IFFs et les GMs des fichier $F_1$ et $F_2$, nous allons appliquer l'algorithme \textsl{UFCIGs pruning}.
En effet, pour générer et affecter les GMs aux IFFs adéquats du fichier Fres, un traitement doit être effectué comme suit ( :
Un motif(Itemset) fréquent est dit fermé s’il ne possède aucun sur-motif qui a le même support. 
\begin{itemize}
    \item $\textsl{EMPYSET} \subset  \textsl{"1 3 5 7 9 13 17"}$, et Supp(EMPTYSET)= Supp(1 3 5 7 9 13 17)=12. Donc "EMPTYSET" n'est plus un itemset fermés fréquent et il sera élagué.
    \item $\textsl{"11"}  \subset \textsl{"1 3 5 7 9 11 13 15 17"}$, et Supp(11)= Supp(1 3 5 7 9 11 13 15 17)=9. Donc "11" n'est plus un itemset fermés fréquent et il sera élagué.
    \item  $\textsl{"19"} \subset \textsl{" 1 3 5 7 9 13 17 19"}$,
    et Supp(19)= Supp(1 3 5 7 9 13 17 19)=9. Donc "19" n'est plus un itemset fermés fréquent et il sera élagué.
    \item $  \textsl{"15"} \subset \textsl{1 3 5 7 9 13 15 17}$,
        et Supp(15)= Supp(1 3 5 7 9 13 15 17 )=10. Donc "15" n'est plus un itemset fermés fréquent et il sera élagué.
\end{itemize}

 \begin{figure}[H]
\centering
\includegraphics[scale=1.2]{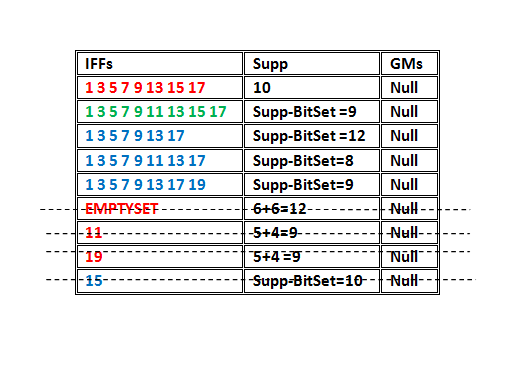}
\caption{ partie d'élagage}
\label{fig444444}
\end{figure}

 Pour affecter les GMs aux IFFs restants dans Fres 
     \begin{itemize}
         \item {IFF "1 3 5 7 9 13 17" son GM ="EMPTYSET".\\
        Supp(1 3 5 7 9 13 17) = Supp(EMPTYSET) = 12, et "EMPTYSET"= $plus\_ petit$.}
        \item{IFF"1 3 5 7 9 11 13 15 17"} son GM= "11". \\
        Supp (1 3 5 7 9 11 13 15 17)=Supp(11), et "11"
         \item{IFF "1 3 5 7 9 13 15 17" son GM = 15.\\
         Supp(1 3 5 7 9 13 15 17) = Supp(15) = 9, et "15"=$plus\_petit$}
         \item{IFF "1 3 5 7 9 11 13 17" son GM= 11 15}.\\
         Supp( 1 3 5 7 9 11 13 17 )= Supp(11 15) = 8 et "11 15" =$plus\_petit$
         \item {IFF"1 3 5 7 9 13 17 19" son GM= "19"}.\\
         Supp ( 1 3 5 7 9 13 17 19)=Supp(19), et "19" =$plus\_petit$
     \end{itemize}

  \begin{figure}[H]
\centering
\includegraphics[scale=1.0]{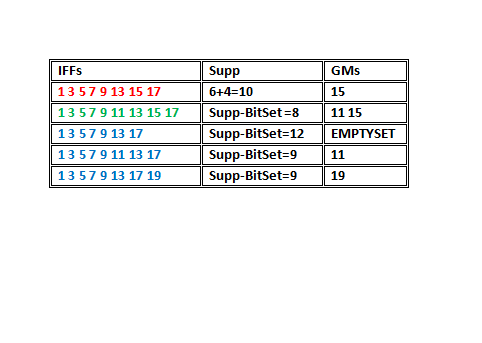}
\caption{ Fichier résultat Fres de l'algorithme 4}
\label{figgm}
\end{figure}
La figure \ref{figgm}  résultat final Fres de notre algorithme UFCIGs pruning  après la phase d'élagage et d'affectation des GMs aux IFFs correspondants.

\begin{figure}[H]
\centering
\includegraphics[scale=0.8]{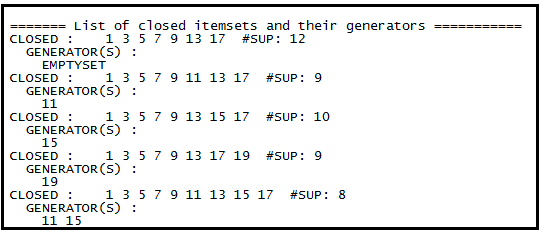}
\caption{ Extraction séquentielle des IFFs,GMs.  }
\label{figseq}
\end{figure}

La figure \ref{figseq} montre le résultat d'extraction séquentielle des IFFs avec leurs GMs t sur la totalité de la base, c'est à dire sans partitionner le contexte initial.
Donc Comparons ce fichier avec notre fichier résultat Fres, nous remarquons que nous avons trouvé les mêmes IFFs ainsi que leurs GMS associés et les mêmes valeurs de supports.

\section{Conclusion}
Dans ce chapitre, nous avons proposé une nouvelle approche permettant l'extraction  des itemsets fermés fréquents , ainsi que leurs générateurs minimaux associés,  dans une base transactionnelle, en les mettant à jour selon  la stratégie "diviser pou régner". Dans le chapitre suivant, nous allons tester les performance de notre algorithme \textsl{UFCIGs-DAC} sur des bases de tests.

\newpage
 \chapter{Étude Expérimentale}
 \section{Introduction}
 
 Dans le chapitre précédent, nous avons introduit un nouvel algorithme, appelé \textsl{UFCIGs-DAC} dédié à l'extraction des motifs fermés fréquents avec leurs générateurs minimaux associés. En effet, notre algorithme, opère en trois étapes successives:
 \begin{enumerate}
     \item {Partitionner la base de transactions initiale. }
     \item {Appliquer sur chaque partition un algorithme d'extraction des motifs fermés fréquents avec leurs générateurs minimaux (par exemple  ZART\cite{szathmary2006zart}).}
     \item{La mise à jour des motifs fermés fréquents ainsi que leurs générateurs minimaux.}
 \end{enumerate}
Dans ce chapitre, nous allons discuter les résultats des expérimentations que nous avons réalisé avec notre approche sur plusieurs bases de tests  pour évoluer ses performances. Dans un premier lieu, nous allons présenter l'environnement de l'évaluation de notre approche. Dans un deuxième lieu, nous allons présenter les  bases "benchmark". Ensuite, nous comparerons les performances de notre algorithme \textsl{UFCIGs-DAC}  à l’algorithme séquentiel  ZART.

\section{Environnement d'expérimentation}
Dans cette section, nous allons commencer tout d'abord par la présentation de l'environnement expérimental sur lequel nous avons travaillé pour évaluer et tester notre  approche de mise à jour des motifs fermés fréquents et  leurs générateurs minimaux   .
 \subsection*{Environnement matériel et logiciel}
 Toutes les expérimentations ont été réalisées sur PC muni d'un processeur Intel Core i3 ayant une fréquence d'horloge de 2.10 GHz et 4 Go de mémoire tournant sur la plate-forme Windows 7.
 Afin de mener une étude comparative avec les approches d'extractions des itemsets fermé fréquents dans le chapitre 2, nous avons implémenté notre algorithme UFCGs-DAC, en java.
 \section{Bases de tests}
 Dans cette section, nous présentons les résultats de l’étude expérimentale que nous avons réalisée sur les bases “benchmark” MUSHROOMS, CHESS, Retail, et Foodmart.
 Typiquement, les bases de transactions réélles sont très denses et elles produisent un nombre important d'itemsets fréquents de taille assez large, et ce même pour
des valeurs élevées de support. Les bases de transactions synthétiques limitent les transactions d'un environnement de ventes au détail. Habituellement, les bases
de transactions synthétiques sont plus éparses comparées aux bases réelles. 
La table \ref{tabdes} énumère les caractéristiques des différentes bases que nous avons utilisé pour nos tests.

\begin{center}

    \begin{tabular}{|p{4cm}||p{4cm}||p{3cm}||p{5cm}|}
        \hline \hline     
  Bases  &Type du contexte&   Nombre d'items & Nombre de transactions\\
         \hline    
Mushrooms &Dense& 119 & 8124\\  
\hline
Chess & Dense& 75 & 3196\\
\hline\hline
Retail & Épars & 10107 &19 211\\
\hline 
Foodmart&Épars&1559&4141 \\
\hline \hline
\end{tabular} 
\captionof{table}{ Caractéristiques des bases transactionnelles de test}
\label{tabdes}
    \end{center}

\section{Résultats expérimentaux}
Nous présentons les résultats obtenus suite aux différentes expérimentations
réalisées dans l'objectif de comparer les performances de UFCIGs-DAC. Tout d'abord,nous avons partitionner les contextes de tests en deux sous-contextes (partitions $P_1$, $P_2)$, dont les caractéristiques sont résumées par le tableau \ref{tabpart}. Puis nous avons extrait les partitions $P_1$ et $P_2$ simultanément en utilisant les Threads. Enfin, nous avons mis à jour l'ensemble des motifs fermés fréquents et leurs générateurs minimaux extrait par ZART.

\begin{center}

    \begin{tabular}{|p{3cm}||p{2cm}||p{2cm}||p{2cm}||p{2cm}|}
        \hline \hline     
  Bases &$\left | \mathit{T} \right |P1  $ &$ \left | \mathit{T} \right |$ P2 & $\left | \mathit{I} \right |P1  $ &   $\left | \mathit{I} \right |P2  $ \\
         \hline    
Mushrooms & 4068 & 4 068& 88& 110\\  
\hline
Chess & 1 597 & 1 599&67&75 \\
\hline\hline
 
Retail &9 474 & 9 737&14 027& 12 191\\
\hline 
Foodmart& 2 071& 2 070&1 551& 1 553\\
\hline \hline
\end{tabular} 
\captionof{table}{ Caractéristiques des Partitions}
\label{tabpart}
    \end{center}
    
Dans ce qui suit, nous allons évaluer les performances de notre algorithme UFCIGs-DAC par rapport à l'algorithme ZART selon deux parties distinguées :
\begin{itemize}
    \item {Temps d'exécution de UFCIGs-DAC versus ZART pour les bases denses et éparses.}
     \item {Nombre des itemsets fermés fréquent extraits par UFCIGs-DAC par rapport à ZART pour les bases denses et éparses.}
\end{itemize}

\subsection{Temps d'exécution de UFCIGs-DAC versus ZART}
\subsubsection{Expérimentations sur les contextes denses}
\begin{figure}[H]
\centering
\includegraphics[scale=0.6]{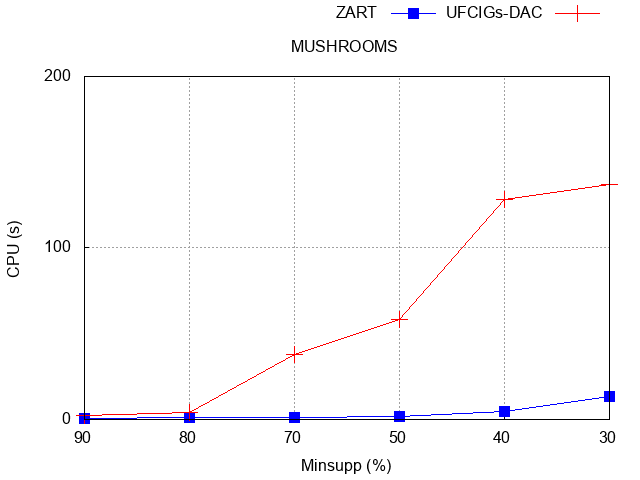}
\caption{Les performances de UFCIGs-DAC  versus ZART dans la base \textbf{\textrm{MUSHROOMS}} }
\label{figtempsmushrooms}
\end{figure}

\begin{figure}[H]
\centering
\includegraphics[scale=0.6]{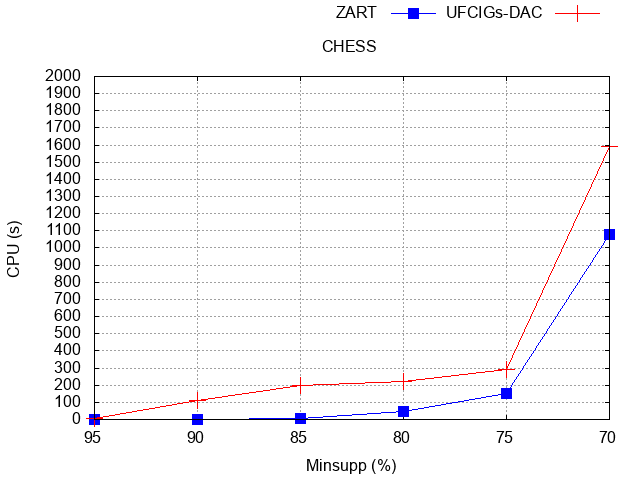}
\caption{Les performances de UFCIGs-DAC  versus ZART dans la base \textbf{\textrm{CHESS}} }
\label{figtempschess}
\end{figure}

Les temps d’exécution de l’algorithme UFCIGs-DAC comparés respectivement à l'algorithme séquentiel ZART sur  les contextes qui sont présentés par les Figures \ref{figtempsmushrooms}, \ref{figtempschess}.

\begin{itemize}
    \item {\textbf{MUSHROOMS :}  pour cette base \textit{ZART} fait mieux que \textit{UFCIGs-DAC} avec un seuil de support minimum très petit.Les performances de UFCIGs-DAC se dégradent considérablement étant donné qu’ils effectuent des intersectionsn sur un grand nombre d’objets de taille élevée à partir de valeur de minsupp = $70\%$. En effet les motifs fermés fréquents sont longs et nombreux, et c'est du au fait que les partitions $P_1$ et $P_2$ partagent de nombreux items en commun).Donc, pour passer à la phase de mise à jour,  il faudra attendre ZART très long  finisse d’extraire les itemsets fermés fréquents avec leurs générateurs minimaux  localement.
 }
    \item{\textbf{CHESS :}  pour cette base,bien qu'il ait eu un partitionnement de la base,UFCIGs-DAC réalise des temps d’exécution beaucoup moins importants dans la partition $P_2$ que ceux réalisés dans la partitions $P_1$. En effet, la performance de UFCIGs-DAC dépend fortement de la fouille de l'algorithme ZART dans les partitions. C’est du au fait que la partition  $P_1$contient plusieurs transactions similaires alors le nombre de motifs qui sont fréquents localement dans cette partition est élevé. Dans ce cas, le temps d’exécution de ZART sur $P_1$ serait élevé  ce qui impacte la performance globale de UFCIGs-DAC.}
    
\end{itemize}


\subsubsection{Expérimentations sur les contextes épars}
\begin{figure}[H]
\centering
\includegraphics[scale=0.8]{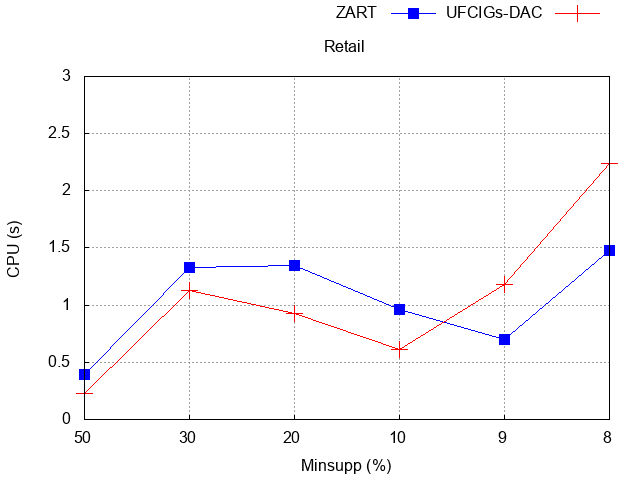}
\caption{Les performances de UFCIGs-DAC  versus ZART dans la base \textbf{\textrm{Retail}} }
\label{figtempsretail}
\end{figure}

\begin{figure}[H]
\centering
\includegraphics[scale=0.8]{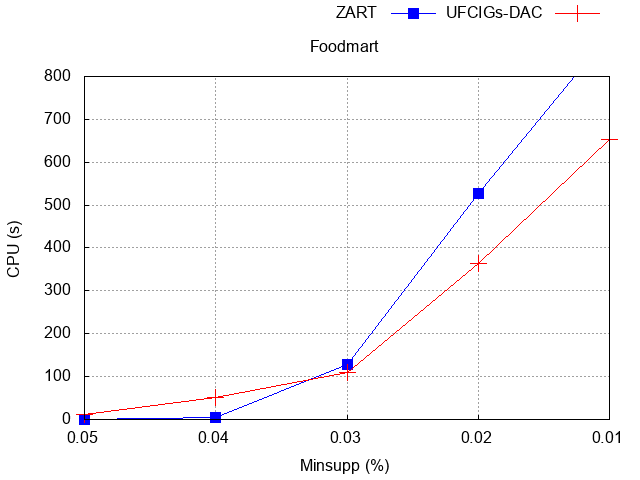}
\caption{Les performances de UFCIGs-DAC  versus ZART dans la base \textbf{\textrm{Foodmart}} }
\label{figtempsfoodmart}
\end{figure}

Les temps d’exécution de l’algorithme UFCIGs-DAC comparés respectivement à l'algorithme séquentiel ZART sur  les contextes épars qui sont présentés par les Figures \ref{figtempsretail} ,\ref{figtempsfoodmart}.




\begin{itemize}

    \item{\textbf{Retail:} pour cette base, les performances  de \textit{UFCIGs-DAC } sont meilleurs que  \textit{ZART} pour les valeurs de minsupp (\textbf {$50  \%,  30\%,  20\%,  et  10\%$}). Alors que c’est l’inverse pour les valeurs de minsupp (\textbf{$2\% et 1\%$}). Ceci peut être expliqué par le fait que UFCIGs-DAC est pénalisé par le coût le calcul des supports Bitset-Rpr, c'est à dire, pour les valeurs de minsupp (\textbf{$2\% et 1\%$}), les itemsets fermés fréquents se trouvant dans l'une des partitions sont plus nombreux que les itemsets fermés fréquents se trouvant dans les deux partitions à la fois.   }
    \item {\textbf{Foodmart :}  dans le cas du contexte Foodmart, les performance de \textit{UFCIGs-DAC } sont largement meilleurs  que celles de \textit{ZART} pour des valeurs de minsupp inférieures ou égales à \textbf{$0,4 \%$}. Les performances réalisées peuvent être expliquées la taille moyenne des motifs fermés fréquents relativement petite sur lesquelles ils exécutent des intersections. De surcroît,
    la majorité des  iemsets fermés fréquents extraits de $P_1$ égales aux iemsets fermés fréquents extraits de $P_2$ .

}
\end{itemize}

\subsection{Nombre des itemsets fermés fréquents extraits}

 Une étude statique qui permet de déterminer la fréquence en pourcentage (\%) de l'algorithme UFCIGs-DAC, c'est à dire, nous allons indiquer le nombre des IFFs retournés par UFCIGs-DAC par rapport à la totalité (le nombre des IFFs retourné par l'algorithme ZART).

\subsubsection{Expérimentations sur les contextes denses}
\begin{figure}[H]
\centering
\includegraphics[scale=0.6]{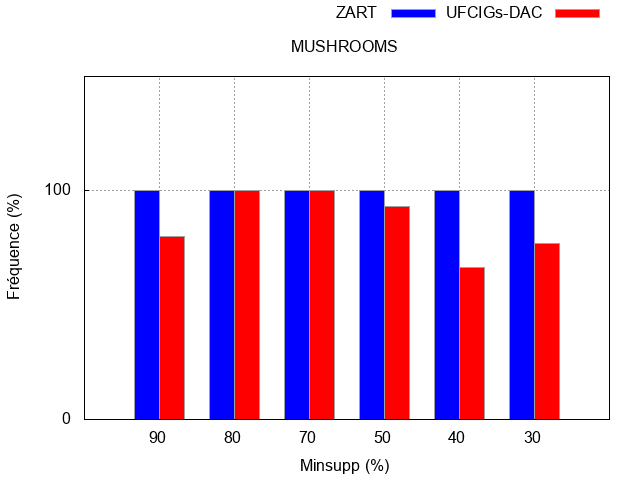}
\caption{Nombre des IFFs extrait par UFCIGs-DAC par rapport à ZART.}
\label{figiff}
\end{figure}

Le nombre des itemsets fermés fréquents extraits par  l’algorithme UFCIGs-DAC  comparés respectivement à l'algorithme ZART  sur  les contextes qui sont présentés par la Figure \ref{figiff} et
le Tableau \ref{tabiff}. 

\newpage

 \begin{itemize}
     \item {\textbf{MUSHROOMS:} dans cette base, a réussi à extriare  toue les motifs fermés fréquents extraits par Zart pour les valeurs de mmissup = $80\%$ et $70\% $. Cependant, pour les autres valeurs de minsupports UFCIGs-DAC  a pratiquement donné les mêmes résultats que nous jugeons satisfaisants.}

     \item{\textbf{CHESS:} dans cette base, 
     comparé à l'algorithme séquentiel ZART, UFCIGs-DAC n'a pas extrait tous les motifs fermés fréquents quelles que soient les valeurs de minsupp, mais l’ensemble des motifs fermés fréquents extrait, présente une fréquence d’apparition jugée satisfaisante. De son coté, UFCIGs-DAC pose un problème de perte d'information dans des certains valeur de support miniumum par rapport à l'algorithme Zart. En outre, ceci est expliqué par la décomposition des contextes qui peut provoquer la disparition des certaines motifs fermés fréquents.
     }
    
 \end{itemize}

\begin{figure}[H]
\centering
\includegraphics[scale=0.6]{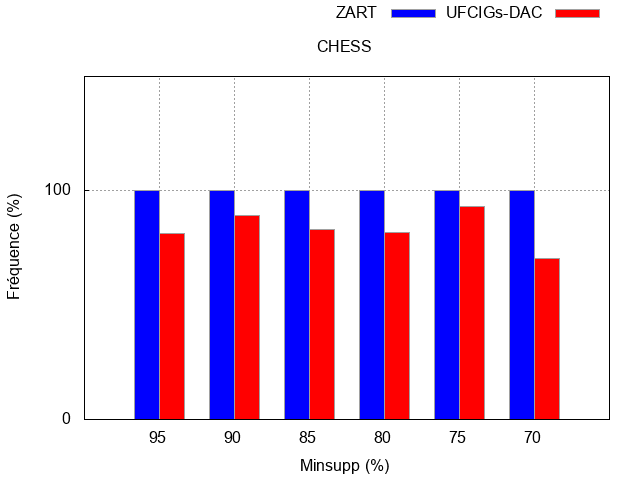}
\caption{Nombre des IFFs extraits par UFCIGs-DAC par rapport à ZART.}
\label{figifff}
\end{figure}

\subsubsection{Expérimentations sur les contextes épars}
Le nombre des itemsets fermés fréquents extraits par  l’algorithme UFCIGs-DAC  comparés respectivement à l'algorithme ZART  sur  les contextes qui sont présentés par la Figure \ref{figiff} et
le Tableau \ref{tabiff}.

\begin{itemize}
    \item{\textbf{Retail:} dans le cas du contexte Retail, \textit{UFCIGs-DAC} bénéficie du partitionnement de ce contexte, et il a réussi à extraire pratiquement les mêmes résultats que \textit{ZART}. En effet le type de la base Retail qui est considérée comme une base éparse, c'est à dire, elle produit un nombre d'itmsets fermés fréquent de taille assez petit par rapport aux bases denses. }

     \item{\textbf{Foodmart:}  dans cette base, comparé à \textit{ZART}, notre algorithme a extrait tous les motifs fermés fréquents quelle que soit la valeur de minsupp.
        }
\end{itemize}

\begin{figure}[H]
\centering
\includegraphics[scale=0.6]{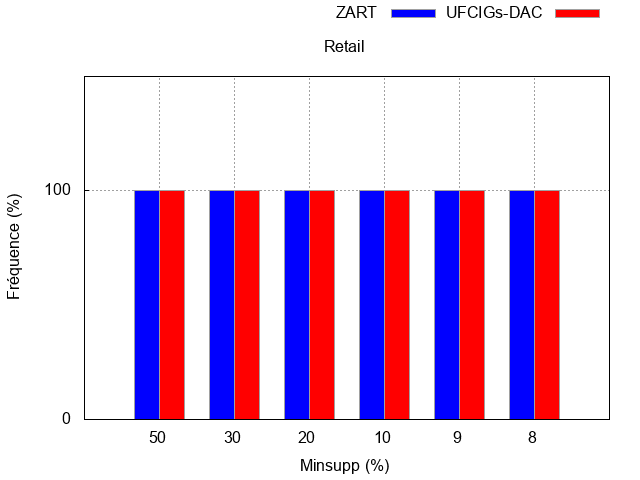}
\caption{Nombre des IFFs extrait par UFCIGs-DAC par rapport à ZART.}
\label{figiffff}
\end{figure}

\begin{figure}[H]
\centering
\includegraphics[scale=0.6]{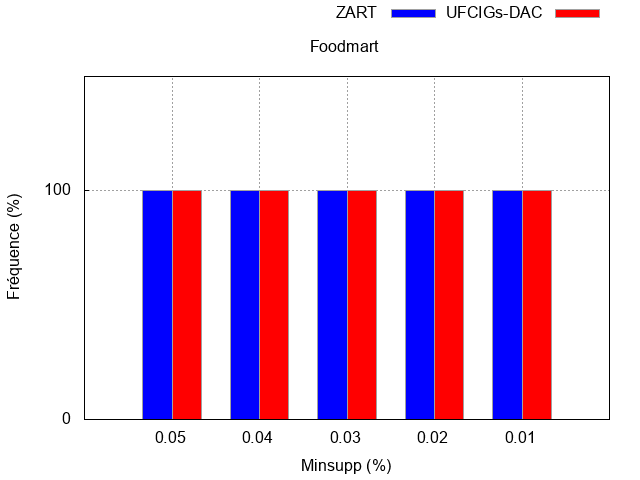}
\caption{Nombre des IFFs extraits par UFCIGs-DAC par rapport à ZART.}
\label{figifffff}
\end{figure}

\section{Interprétation des résultats}

Dans le cas des contextes épars,les performances de UFCIGs-DAC s'avèrent largement meilleures que celles de Zart. En effet, les motifs fermés fréquents extraits des contextes épars sont aussi les générateurs minimaux, tandis que dans les contextes denses il y'a des plusieurs motifs longs même pour des valeurs de supports élevées. Dans ce cas, UFCIGs-DAC faudra attendre très long l'alogrithme d'extraction spécifique (ZART)  finisse d’extraire les itemsets localement. C'est pour cela, Zart a un écart de temps d'exécution  meilleur que UFCIGs-DAC dans les bases denses. De son coté, UFCIGs-DAC pose un problème de perte d'information dans des certains valeur de support miniumum par rapport à l'algorithme Zart. En outre, ceci est expliqué par la décomposition des contextes qui peut provoquer la disparition des certaines motifs fermés fréquents. Par contre, dans les bases éparses, le temps d'exécution de l'algorithme UFCIGs-DAC commence à être  distinguable. Les performances réalisées peuvent être expliquées par la taille moyenne des motifs fermés fréquents relativement petite sur lesquelles ils exécutent des intersections. Ainsi, notre algorithme a réussi à extraire pratiquement les mêmes résultats que ZART.
    

 \section{Conclusion}
 Dans ce chapitre, nous avons mené une étude expérimentale de l'algorithme UFCIGs-DAC sur des bases "benchmark" communément utilisées. Nous avons prouvé expérimentalement que nous pouvons réduire le temps d'exécution   des motifs fermés fréquents et leurs générateurs minimaux associés .
 \chapter*{Conclusion générale }
 \markboth{CONCLUSION GÉNÉRALE}{}
\addcontentsline{toc}{chapter}{Conclusion générale}
 Dans ce mémoire, nous nous sommes intéressés à la fouille des motifs fermés fréquents dans les baeses des transactions. À cet égard, nous avons proposé, dans ce mémoire, une nouvelle approche permettant l'extraction des itemsets fermés fréquents. En effet,  nous avons entamé ce mémoire par la présentation des notions préliminaires relatives aux motifs fréquents et aux motifs fermés fréquents. Nous avons tout de même décrit les notions offertes par le cadre  de l’analyse des concepts formels ACF. 
Ensuite, nous avons étudié dans le deuxième chapitre, les différentes approches de la littérature traitant de l'extraction séquentielle des motifs fréquents, des motifs fermés fréquents ainsi que les approches parallèles de fouille. Egalement, nous avons mené une étude critique des principaux algorithmes d’extraction dans les bases massives en se basant sur une stratégie de partitionnement « Diviser pour régner » du contexte de données. 
Notre nouvelle approche appelé  UFCIGs-DAC a été conçu et implémenté afin de réaliser la fouille dans les bases de tests. La principale originalité de notre approche est l’exploration simultanée de l’espace de recherche en mettant à jour les motifs fermés fréquents et les générateurs minimaux.
De plus, notre approche pourrait être adaptée à tout algorithme d'extraction des motifs fermés fréquents avec leurs générateurs minimaux.

Dans le bus d'améliorer l'exploitation et la flexibilitéion notre approches, en voici quelques perspectives  que nous jugeons intéressantes:
\begin{itemize}
    \item {Application  d’une stratégie de partitionnement non-aléatoire du contexte. Autrement dit, en prenant en considération l’équité du nombre d’items dans les différentes partitions ainsi que leur répartition. Par exemple, quand une partition contient plusieurs transactions similaires (elles partagent de nombreux items en commun), alors le nombre de motifs qui sont fréquents localement dans cette partition est élevé et par la suite le nombre des motifs fermés fréquents.  }
    \item{ Adapter notre approche pour la fouille des "Big Data"   dans les environnements  distribués ( Hadoop \cite{holmes2012hadoop}, Spark \cite{zaharia2010spark}).}
    \item{Entamer l’étape d’extraction des règles associatives \cite{agrawal1993mining},\cite{bouzouita2006garc}. Une règle d’association de la forme $X \rightarrow Y$, où X et Y sont des motifs disjoints ($X \cap Y =  \phi $), appelées respectivement la prémisse et la conclusion de la règle. Cette règle est traduite par " si X, alors Y ". Notons que X c’est l’Itemset fermé fréquent et Y est le générateur minimal associé au fermé X \cite{gasmi2007extraction},\cite{bouker2012ranking}.}
\end{itemize}

\nocite{hamrouni}
\nocite{draheim2017generalized}
\nocite{draheim2017semantics}
\nocite{auer2009extending}
\nocite{draheim2010service}
\nocite{atkinson2010typed}
\nocite{mouakher2016qualitycover}
\nocite{mouakher2019efficient}

\bibliographystyle{plain}

\end{document}